# A Heuristic Search Approach to Planning with Continuous Resources in Stochastic Domains

**Nicolas Meuleau** NICOLAS.F.MEULEAU@NASA.GOV
*NASA Ames Research Center*
*Mail Stop 269-3*
*Moffet Field, CA 94035-1000, USA*

**Emmanuel Benazera** EBENAZER@LAAS.FR
*LAAS-CNRS, Université de Toulouse*
*7, av. du Colonel Roche*
*31077 Toulouse Cedex 4, France*

**Ronen I. Brafman** BRAFMAN@CS.BGU.AC.IL
*Department of Computer Science*
*Ben-Gurion University*
*Beer-Sheva 84105, Israel*

**Eric A. Hansen** HANSEN@CSE.MSSTATE.EDU
*Department of Computer Science and Engineering*
*Mississippi State University*
*Mississippi State, MS 39762, USA*

**Mausam** MAUSAM@CS.WASHINGTON.EDU
*Department of Computer Science and Engineering*
*University of Washington*
*Seattle, WA 981952350, USA*

## Abstract

We consider the problem of optimal planning in stochastic domains with resource constraints, where the resources are continuous and the choice of action at each step depends on resource availability. We introduce the HAO* algorithm, a generalization of the AO* algorithm that performs search in a hybrid state space that is modeled using both discrete and continuous state variables, where the continuous variables represent monotonic resources. Like other heuristic search algorithms, HAO* leverages knowledge of the start state and an admissible heuristic to focus computational effort on those parts of the state space that could be reached from the start state by following an optimal policy. We show that this approach is especially effective when resource constraints limit how much of the state space is reachable. Experimental results demonstrate its effectiveness in the domain that motivates our research: automated planning for planetary exploration rovers.

## 1. Introduction

Many NASA planetary exploration missions rely on rovers – mobile robots that carry a suite of scientific instruments for use in characterizing planetary surfaces and transmitting information back to Earth. Because of difficulties in communicating with devices on distant planets, direct human control of rovers by tele-operation is infeasible, and rovers must be able to act autonomously for substantial periods of time. For example, the Mars Exploration Rovers (MER), *aka*, Spirit and Opportunity, are designed to communicate with the ground only twice per Martian day.

Autonomous control of planetary exploration rovers presents many challenges for research in automated planning. Progress has been made in meeting some of these challenges. For example, the planning software developed for the Mars Sojourner and MER rovers has contributed significantly to the success of these missions (Bresina, Jonsson, Morris, & Rajan, 2005). But many important challenges must still be addressed to achieve the more ambitious goals of future missions (Bresina, Dearden, Meuleau, Ramakrishnan, Smith, & Washington, 2002).

Among these challenges is the problem of plan execution in uncertain environments. On planetary surfaces such as Mars, there is uncertainty about the terrain, meteorological conditions, and the state of the rover itself (position, battery charge, solar panels, component wear, etc.) In turn, this leads to uncertainty about the outcome of the rover's actions. Much of this uncertainty is about resource consumption. For example, factors such as slope and terrain affect speed of movement and rate of power consumption, making it difficult to predict with certainty how long it will take for a rover to travel between two points, or how much power it will consume in doing so. Because of limits on critical resources such as time and battery power, rover plans are currently very conservative and based on worst-case estimates of time and resource usage. In addition, instructions sent to planetary rovers are in the form of a sequential plan for attaining a single goal (e.g., photographing an interesting rock). If an action has an unintended outcome that causes a plan to fail, the rover stops and waits for further instructions; it makes no attempt to recover or achieve an alternative goal. This can result in under-utilized resources and missed science opportunities.

Over the past decade, there has been a great deal of research on how to generate conditional plans in domains with uncertain action outcomes. Much of this work is formalized in the framework of Markov decision processes (Puterman, 1994; Boutilier, Dean, & Hanks, 1999). However, as Bresina et al. (2002) point out, important aspects of the rover planning problem are not adequately handled by traditional planning algorithms, including algorithms for Markov decision processes. In particular, most traditional planners assume a discrete state space and a small discrete number of action outcomes. But in automated planning for planetary exploration rovers, critical resources such as time and battery power are continuous, and most of the uncertainty in the domain results from the effect of actions on these variables. This requires a conditional planner that can branch not only on discrete action outcomes, but on the availability of continuous resources, and such a planner must be able to reason about continuous as well as discrete state variables.

Closely related to the challenges of uncertain plan execution and continuous resources is the challenge of over-subscription planning. The rovers of future missions will have much improved capabilities. Whereas the current MER rovers require an average of three days to visit a single rock, progress in areas such as automatic instrument placement will allow rovers to visit multiple rocks and perform a large number of scientific observations in a single communication cycle (Pedersen, Smith, Deans, Sargent, Kunz, Lees, & Rajagopalan, 2005). Moreover, communication cycles will lengthen substantially in more distant missions to the moons of Jupiter and Saturn, requiring longer periods of autonomous behavior. As a result, space scientists of future missions are expected to specify a large number of science goals at once, and often this will present what is known as an *over-subscription planning problem.* This refers to a problem in which it is infeasible to achieve all goals, and the objective is to achieve the best subset of goals within resource constraints (Smith, 2004). In the case of the rover, there will be multiple locations the rover could reach, and many experiments the rover could conduct, most combinations of which are infeasible due to resource constraints. The planner must select a feasible subset of these that maximizes expected science return. When action outcomes (including resource consumption) are stochastic, a plan that maximizes expected science return will be a conditional plan that prescribes different courses of action based on the results of previous actions, including resource availability.

In this paper, we present an implemented planning algorithm that handles all of these problems together: uncertain action outcomes, limited continuous resources, and over-subscription planning. We formalize the rover planning problem as a *hybrid-state Markov decision process*, that is, a Markov decision process (MDP) with both discrete and continuous state variables, and we use the continuous variables to represent resources. The planning algorithm we introduce is a heuristic search algorithm called HAO*, for Hybrid-state AO*. It is a generalization of the classic AO* heuristic search algorithm (Nilsson, 1980; Pearl, 1984). Whereas AO* searches in discrete state spaces, HAO* solves planning problems in hybrid domains with both discrete and continuous state variables. To handle hybrid domains, HAO* builds on earlier work on dynamic programming algorithms for continuous and hybrid-state MDPs, in particular, the work of Feng et al. (2004).

Generalizing AND/OR graph search for hybrid state spaces poses a complex challenge, and we only consider a special case of the problem. In particular, continuous variables are used to represent monotonic resources. The search is for the best conditional plan that allows branching not only on the values of the discrete variables, but on the availability of these resources, and does not violate a resource constraint.

It is well-known that heuristic search can be more efficient than dynamic programming because it uses reachability analysis guided by a heuristic to focus computation on the relevant parts of the state space. We show that for problems with resource constraints, including over-subscription planning problems, heuristic search is especially effective because resource constraints can significantly limit reachability. Unlike dynamic programming, a systematic forward search algorithm such as AO* keeps track of the trajectory from the start state to each reachable state, and thus it can check whether the trajectory is feasible or violates a resource constraint. By pruning infeasible trajectories, a heuristic search algorithm can dramatically reduce the number of states that must be considered to find an optimal policy. This is particularly important in our domain where the discrete state space is huge (exponential in the number of goals), and yet the portion reachable from any initial state is relatively small, due to resource constraints.

## 2. Problem Formulation and Background

We start with a formal definition of the planning problem we are tackling. It is a special case of a hybrid-state Markov decision process, and so we first define this model. Then we discuss how to include resource constraints and formalize over-subscription planning in this model. Finally we review a class of dynamic programming algorithms for solving hybrid-state MDPs, since some of these algorithmic techniques will be incorporated in the heuristic search algorithm we develop in Section 3.

### 2.1 Hybrid-State Markov Decision Process

A hybrid-state Markov decision process, or *hybrid-state MDP*, is a factored Markov decision process that has both discrete and continuous state variables. We define it as a tuple $(N, \mathbf{X}, A, P, R)$, where $N$ is a discrete state variable, $\mathbf{X} = \{X_1, X_2, ..., X_d\}$ is a set of continuous state variables, $A$ is a set of actions, $P$ is a stochastic state transition model, and $R$ is a reward function. We describe these elements in more detail below. A hybrid-state MDP is sometimes referred to as simply a hybrid MDP. The term "hybrid" does not refer to the dynamics of the model, which are discrete. Another term for a hybrid-state MDP, which originates in the Markov chain literature, is a *general-state MDP*.

Although a hybrid-state MDP can have multiple discrete variables, this plays no role in the algorithms described in this paper, and so, for notational convenience, we model the discrete component of the state space as a single variable $N$. Our focus is on the continuous component. We assume the domain of each continuous variable $X_i \in \mathbf{X}$ is a closed interval of the real line, and so $\mathbf{X} = \bigotimes_i X_i$ is the hypercube over which the continuous variables are defined. The state set $S$ of a hybrid-state MDP is the set of all possible assignments of values to the state variables. In particular, a *hybrid state* $s \in S$ is a pair $(n, \mathbf{x})$ where $n \in N$ is the value of the discrete variable, and $\mathbf{x} = (x_i)$ is a vector of values of the continuous variables.

State transitions occur as a result of *actions*, and the process evolves according to Markovian *state transition probabilities* $\Pr(s' \mid s, a)$, where $s = (n, \mathbf{x})$ denotes the state before action $a$ and $s' = (n', \mathbf{x}')$ denotes the state after action $a$, also called the arrival state. These probabilities can be decomposed into:

- the discrete marginals $\Pr(n'|n, \mathbf{x}, a)$. For all $(n, \mathbf{x}, a)$, $\sum_{n' \in N} \Pr(n'|n, \mathbf{x}, a) = 1$;
- the continuous conditionals $\Pr(\mathbf{x}'|n, \mathbf{x}, a, n')$. For all $(n, \mathbf{x}, a, n')$, $\int_{x' \in \mathbf{X}} \Pr(\mathbf{x}'|n, \mathbf{x}, a, n') d\mathbf{x}' = 1$.

We assume the *reward* associated with a transition is a function of the arrival state only, and let $R_n(\mathbf{x})$ denote the *reward* associated with a transition to state $(n, \mathbf{x})$. More complex dependencies are possible, but this is sufficient for the goal-based domain models we consider in this paper.

### 2.2 Resource Constraints and Over-Subscription Planning

To model the rover planning problem, we consider a special type of MDP in which the objective is to optimize expected cumulative reward *subject to resource constraints.* We make the following assumptions:

- there is an initial allocation of one or more non-replenishable resources,
- each action has some minimum positive consumption of at least one resource, and
- once resources are exhausted, no further action can be taken.

One way to model an MDP with resource constraints is to formulate it as a *constrained MDP*, a model that has been widely studied in the operations research community (Altman, 1999). In this model, each action $a$ incurs a transition-dependent resource cost, $C_a^i(s, s')$, for each resource $i$. Given an initial allocation of resources and an initial state, linear programming is used to find the best feasible policy, which may be a randomized policy. Although a constrained MDP models resource consumption, it does not include resources in the state space. As a result, a policy cannot be conditioned upon resource availability. This is not a problem if resource consumption is either deterministic or unobservable. But it is not a good fit for the rover domain, in which resource consumption is stochastic and observable, and the rover should take different actions depending on current resource availability.

We adopt a different approach to modeling resource constraints in which resources are included in the state description. Although this increases the size of the state space, it allows decisions to be made based on resource availability, and it allows a stochastic model of resource consumption. Since resources in the rover domain are continuous, we use the continuous variables of a hybrid-state MDP to represent resources. Note that the duration of actions is one of the biggest sources of uncertainty in our rover problems, and we model time as one of the continuous resources. Resource constraints are represented in the form of executability constraints on actions, where $A_n(\mathbf{x})$ denotes the set of actions executable in state $(n, \mathbf{x})$. An action cannot be executed in a state that does not satisfy its minimum resource requirements.

Having discussed how to incorporate resource consumption and resource constraints in a hybrid-state MDP, we next discuss how to formalize *over-subscription planning.* In our rover planning problem, scientists provide the planner with a set of "goals" they would like the rover to achieve, where each goal corresponds to a scientific task such as taking a picture of a rock or performing an analysis of a soil sample. The scientists also specify a utility or reward for each goal. Usually only a subset of these goals is feasible under resource constraints, and the problem is to find a feasible plan that maximizes expected utility. Over-subscription planning for planetary exploration rovers has been considered by Smith (2004) and van den Briel et al. (2004) for deterministic domains. We consider over-subscription planning in stochastic domains, especially domains with stochastic resource consumption. This requires construction of conditional plans in which the selection of goals to achieve can change depending on resource availability.

In over-subscription planning, the utility associated with each goal can be achieved only once; no additional utility is achieved for repeating the task. Therefore, the discrete state must include a set of Boolean variables to keep track of the set of goals achieved so far by the rover, with one Boolean variable for each goal. Keeping track of already-achieved goals ensures a Markovian reward structure, since achievement of a goal is rewarded only if it was not achieved in the past. However, it also significantly increases the size of the discrete state space. Maintaining history information to ensure a Markovian reward structure is a simple example of planning with non-Markovian rewards (Thiebaux, Gretton, Slaney, Price, & Kabanza, 2006).

### 2.3 Optimality Equation

The rover planning problem we consider is a special case of a finite-horizon hybrid-state MDP in which termination occurs after an indefinite number of steps. The Bellman optimality equation for this problem takes the following form:

$$\begin{aligned} V_n(\mathbf{x}) &= 0 \text{ when } (n, \mathbf{x}) \text{ is a terminal state; otherwise,} \\ V_n(\mathbf{x}) &= \max_{a \in A_n(\mathbf{x})} \left[ \sum_{n' \in N} \Pr(n' \mid n, \mathbf{x}, a) \int_{\mathbf{x}'} \Pr(\mathbf{x}' \mid n, \mathbf{x}, a, n') \left( R_{n'}(\mathbf{x}') + V_{n'}(\mathbf{x}') \right) d\mathbf{x}' \right] . \end{aligned} \quad (1)$$

We define a *terminal state* as a state in which no actions are eligible to execute, that is, $A_n(\mathbf{x}) = \emptyset$. We use terminal states to model various conditions for plan termination. This includes the situation in which all goals have been achieved; the situation in which resources have been exhausted; and the situation in which an action results in some error condition that requires executing a safe sequence by the rover and terminating plan execution. In addition to terminal states, we assume an explicit *initial state* denoted $(n_0, \mathbf{x}_0)$.

Assuming that resources are limited and non-replenishable, and that every action consumes some resource (and the amount consumed is greater than or equal to some positive quantity $c$), plan execution will terminate after a finite number of steps. The maximum number of steps is bounded by the initial resource allocation divided by $c$, the minimal resource consumption per step. The actual number of steps is usually much less and indefinite, because resource consumption is stochastic and because the choice of action influences resource consumption. Because the number of steps it takes for a plan to terminate is bounded but indefinite, we call this a *bounded-horizon MDP* in contrast to a finite-horizon MDP. However, we note that any bounded-horizon MDP can be converted to a finite-horizon MDP by specifying a horizon that is equal to the maximum number of plan steps, and introducing a no-op action that is taken in any terminal state.

Note that there is usually a difference between the number of plan steps and the time a plan takes to execute. Since we model time as one of the continuous resources, the time it takes to execute a plan step is both state and action dependent, and stochastic.

Given a hybrid-state MDP with a set of terminal states and an initial state $(n_0, \mathbf{x_0})$, the objective is to find a policy, $\pi : (N \times \mathbf{X}) \rightarrow A$, that maximizes expected cumulative reward; specifically, an optimal policy has a value function that satisfies the optimality equation given by Equation (1). In our rover domain, cumulative reward is equal to the sum of rewards for the goals achieved before reaching a terminal state and there is no direct incentive to save resources; an optimal solution saves resources only if this allows achieving more goals. However, our framework is general enough to allow reasoning about both the cost and the availability of resources. For example, an incentive for conserving resources could be modeled by specifying a reward that is proportional to the amount of resources left unused upon entering the terminal state. Note that our framework allows reasoning about both the cost and availability of resources without needing to formulate this as a problem of multi-objective optimization, and we stay in a standard decision-theoretic framework.

### 2.4 Dynamic Programming for Continuous-State and Hybrid-State MDPs

Because the planning problem we consider is a finite-horizon hybrid-state MDP, it can be solved by any algorithm for solving finite-horizon hybrid-state MDPs. Most algorithms for solving hybrid-state (and continuous-state) MDPs rely on some form of approximation. A widely-used approach is to discretize the continuous state space into a finite number of grid points and solve the resulting finite-state MDP using dynamic programming and interpolation (Rust, 1997; Munos & Moore, 2002). Another approach is parametric function approximation; a function associated with the dynamic programming problem – such as the value function or policy function – is approximated by a smooth function of $k$ unknown parameters. In general, parametric function approximation is faster than grid-based approximation, but has the drawback that it may fail to converge, or may converge to an incorrect solution. Parametric function approximation is used by other algorithms for solving continuous-state MDPs besides dynamic programming. Reinforcement learning algorithms use artificial neural networks as function approximators (Bertsekas & Tsitsiklis, 1996). An approach to solving MDPs called approximate linear programming has been extended to allow continuous as well as discrete state variables (Kveton, Hauskrecht, & Guestrin, 2006).

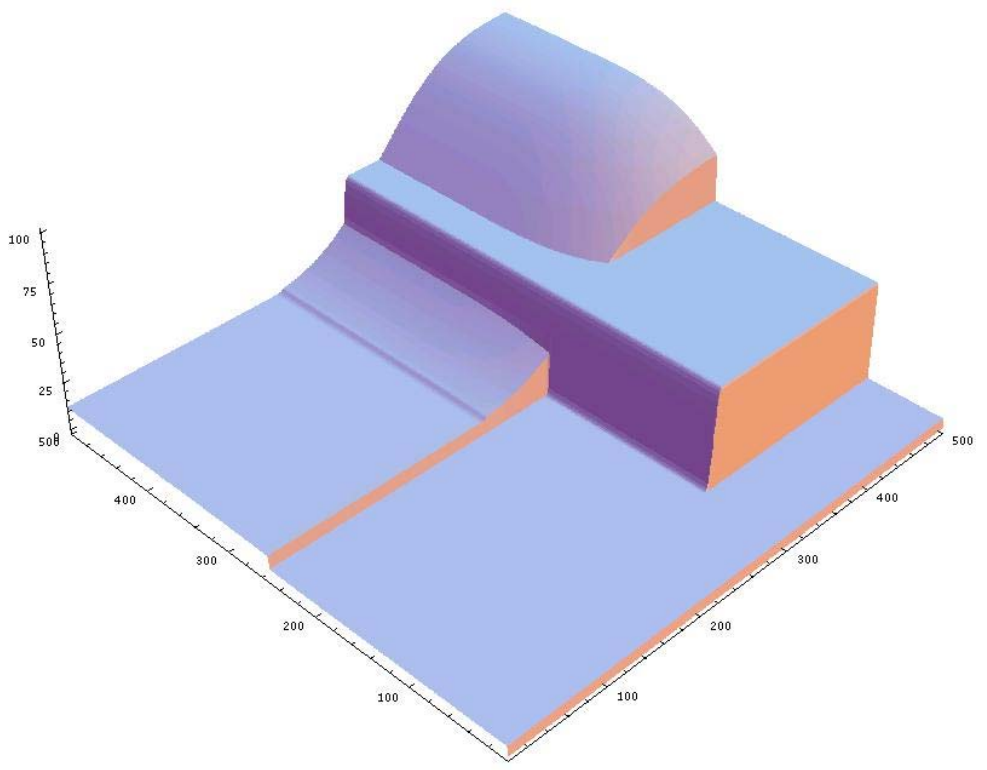

Figure 1: Value function in the initial state of a simple rover problem: optimal expected return as a function of two continuous variables (time and energy remaining).

We review another approach to solving hybrid-state (or continuous-state) MDPs that assumes the problem has special structure that can be exploited by the dynamic programming algorithm. The structure assumed by this approach ensures that the convolution $\int_{\mathbf{x}'} \Pr(\mathbf{x}' \mid n, \mathbf{x}, a, n')(R_{n'}(\mathbf{x}') + V_{n'}(\mathbf{x}'))d\mathbf{x}'$ in Equation (1) can be computed exactly in finite time, and the value function computed by dynamic programming is piecewise-constant or piecewise-linear. The initial idea for this approach is due to the work of Boyan and Littman (2000), who describe a class of MDPs called *time-dependent MDPs*, in which transitions take place along a single, irreversible continuous dimension. They describe a dynamic programming algorithm for computing an exact piecewise-linear value function when the transition probabilities are discrete and rewards are piecewise linear. Feng et al. (2004) extend this approach to continuous state spaces of more than one dimension, and consider MDPs with discrete transition probabilities and two types of reward models: piecewise constant and piecewise linear. Li and Littman (2005) further extend the approach to allow transition probabilities that are piecewise-constant, instead of discrete, although this extension requires some approximation in the dynamic programming algorithm.

The problem structure exploited by these algorithms is characteristic of the Mars rover domain and other over-subscription planning problems. Figure 1 shows the optimal value functions from the initial state of a typical Mars rover problem as a function of two continuous variables: the time and energy remaining (Bresina et al., 2002). The value functions feature a set of humps and plateaus, each of them representing a region of the state space where similar goals are pursued by the optimal policy. The sharpness of a hump or plateau reflects uncertainty about achieving the goal(s). Constraints that impose minimal resource levels before attempting some actions introduce sharp cuts in the regions. Plateau regions where the expected reward is nearly constant represent regions of the state space where the optimal policy is the same, and the probability distribution over future histories induced by this optimal policy is nearly constant.

The structure in such a value function can be exploited by partitioning the continuous state space into a finite number of hyper-rectangular regions. (A region is a hyper-rectangle if it is the Cartesian product of intervals at each dimension.) In each hyper-rectangle, the value function is either constant (for a piecewise-constant function) or linear (for a piecewise-linear function). The resolution of the hyper-rectangular partitioning is adjusted to fit the value function. Large hyper-rectangles are used to represent large plateaus. Small hyper-rectangles are used to represent regions of the state space where a finer discretization of the value function is useful, such as the edges of plateaus and the curved hump where there is more time and energy available. A natural choice of data structures for rectangular partitioning of a continuous space is kd-trees (Friedman, Bentley, & Finkel, 1977), although other choices are possible. Figures 6 and 10 in Section 4.1 show value functions for the initial state of a simple rover planning problem, created by a piecewise-constant partitioning of the continuous state space.

The continuous-state domains of the transition and reward functions are similarly partitioned into hyper-rectangles. The reward function of each action has the same piecewise-constant (or piecewise-linear) representation as the value function. The transition function partitions the state space into regions for which the set of outcomes of an action and the probability distribution over the set of outcomes are identical. Following Boyan and Littman (2000), both relative and absolute transitions are supported. A relative outcome can be viewed as shifting a region by a constant $\delta$. That is, for any two states $x$ and $y$ in the same region, the transition probabilities$Pr(x'|x,a)$ and $Pr(y'|y,a)$ are defined in term of the probability of $\delta$, such that $\delta = (x' - x) = (y' - y)$. An absolute outcome maps all states in a region to a single state. That is, for any two states $x$ and $y$ in the same region, $Pr(x'|x,a) = Pr(x'|y,a)$. We can view a relative outcome as a pair $(\delta, p)$, where $p$ is the probability of that outcome, and we can view an absolute outcome as a pair $(x', p)$. This assumes there is only a finite number of non-zero probabilities, i.e., the probability distribution is discretized, which means that for any state and action, a finite set of states can be reached with non-zero probability. This representation guarantees that a dynamic programming update of a piecewise-constant value function results in another piecewise-constant value function. Feng et al. (2004) show that for such transition functions and for any finite horizon, there exists a partition of the continuous space into hyper-rectangles over which the optimal value function is piecewise constant or linear.

The restriction to discrete transition functions is a strong one, and often means the transition function must be approximated. For example, rover power consumption is normally distributed, and thus must be discretized. (Since the amount of power available must be non-negative, our implementation truncates any negative part of the normal distribution and renormalizes.) Any continuous transition function can be approximated by an appropriately fine discretization, and Feng et al. (2004) argue that this provides an attractive alternative to function approximation approaches in that it approximates the model but then solves the approximate model exactly, rather than finding an approximate value function for the original model. (For this reason, we will sometimes refer to finding optimal policies and value functions, even when the model has been approximated.) To avoid discretizing the transition function, Li and Littman (2005) describe an algorithm that allows piecewise-constant transition functions, in exchange for some approximation in the dynamic programming algorithm. Marecki et al.(2007) describe a different approach to this class of problems in which probability distributions over resource consumptions are represented with phase-type distributions and a dynamic programming algorithm exploits this representation. Although we use the work of Feng et al. (2004) in our implementation, the heuristic search algorithm we develop in the next section could use any of these or some other approach to representing and computing value functions and policies for a hybrid-state MDP.

## 3. Heuristic Search in a Hybrid State Space

In this section, we present the primary contribution of this paper: an approach to solving a special class of hybrid-state MDPs using a novel generalization of the heuristic search algorithm AO*. In particular, we describe a generalization of this algorithm for solving hybrid-state MDPs in which the continuous variables represent monotonic and constrained resources and the acyclic plan found by the search algorithm allows branching on the availability of these resources.

The motivation for using heuristic search is the potentially huge size of the state space, which makes dynamic programming infeasible. One reason for this size is the existence of continuous variables. But even if we only consider the discrete component of the state space, the size of the state space is exponential in the number of discrete variables. As is well-known, AO* can be very effective in solving planning problems that have a large state space because it only considers states that are reachable from an initial state, and it uses an informative heuristic function to focus on states that are reachable in the course of executing a good plan. As a result, AO* can often find an optimal plan by exploring a small fraction of the entire state space.

We begin this section with a review of the standard AO* algorithm. Then we consider how to generalize AO* to search in a hybrid state space and discuss the properties of the generalized algorithm, as well as its most efficient implementations.

### 3.1 AO*

Recall that AO* is an algorithm for AND/OR graph search problems (Nilsson, 1980; Pearl, 1984). Such graphs arise in problems where there are choices (the OR components), and each choice can have multiple consequences (the AND component), as is the case in planning under uncertainty. Hansen and Zilberstein (2001) show how AND/OR graph search techniques can be used in solving MDPs.

Following Nilsson (1980) and Hansen and Zilberstein (2001), we define an AND/OR graph as a hypergraph. Instead of arcs that connect pairs of nodes as in an ordinary graph, a hypergraph has *hyperarcs*, or *k-connectors*, that connect a node to a set of $k$ successor nodes. When an MDP is represented by a hypergraph, each node corresponds to a state; the root node corresponds to the start state, and the leaf nodes correspond to terminal states. Thus we often use the word *state* to refer to the corresponding node in the hypergraph representing an MDP. A $k$-connector corresponds to an action that transforms a state into one of $k$ possible successor states, with a probability attached to each successor such that the probabilities sum to one. In this paper, we assume the AND/OR graph is acyclic, which is consistent with our assumption that the underlying MDP has a bounded-horizon.

In AND/OR graph search, a solution takes the form of an acyclic subgraph called a *solution graph*, which is defined as follows:

- the start node belongs to a solution graph;
- for every non-terminal node in a solution graph, exactly one outgoing $k$-connector (corresponding to an action) is part of the solution graph and each of its successor nodes also belongs to the solution graph;
- every directed path in the solution graph terminates at a terminal node.

A solution graph that maximizes expected cumulative reward is found by solving the following system of equations,

$$V^*(s) = \begin{cases} 0 \text{ if } s \text{ is a terminal state; otherwise,} \\ \max_{a \in A(s)} \left[ \sum_{s' \in S} Pr(s'|s,a) \left( R(s') + V^*(s') \right) \right], \end{cases} \tag{2}$$

where $V^*(s)$ denotes the expected value of an optimal solution for state $s$, and $V^*$ is called the optimal *evaluation function* (or *value function* in MDP terminology). Note that this is identical to the optimality equation for hybrid-state MDPs defined in Equation (1), if the latter is restricted to a discrete state space. In keeping with the convention in the literature on MDPs, we treat this as a value-maximization problem even though AO* is usually formalized as solving a cost-minimization problem.

For state-space search problems that are formalized as AND/OR graphs, an optimal solution graph can be found using the heuristic search algorithm AO* (Nilsson, 1980; Pearl, 1984). Like other heuristic search algorithms, the advantage of AO* over dynamic programming is that it can find an optimal solution for a particular starting state without evaluating all problem states. Therefore, a graph is not usually supplied explicitly to the search algorithm. An implicit graph, $G$, is specified implicitly by a start node or start state $s$ and a successor function that generates the successors states for any state-action pair. The search algorithm constructs an *explicit graph*, $G'$, that initially consists only of the start state. A tip or leaf state of the explicit graph is said to be terminal if it is a goal state (or some other state in which no action can be taken); otherwise, it is said to be nonterminal. A nonterminal tip state can be *expanded* by adding to the explicit graph its outgoing $k$-connectors (one for each action) and any successor states not already in the explicit graph.

AO* solves a state-space search problem by gradually building a solution graph, beginning from the start state. A *partial solution graph* is defined similarly to a solution graph, with the difference that tip states of a partial solution graph may be nonterminal states of the implicit AND/OR graph. A partial solution graph is defined as follows:

- the start state belongs to a partial solution graph;
- for every non-tip state in a partial solution graph, exactly one outgoing $k$-connector (corresponding to an action) is part of the partial solution graph and each of its successor states also belongs to the partial solution graph;
- every directed path in a partial solution graph terminates at a tip state of the explicit graph.

The value of a partial solution graph is defined similarly to the value of a solution graph. The difference is that if a tip state of a partial solution graph is nonterminal, it does not have a value that can be propagated backwards. Instead, we assume there is an admissible heuristic estimate $H(s)$ of the maximal-value solution graph for state $s$. A heuristic evaluation function $H$ is said to be *admissible* if $H(s) \geq V^*(s)$ for every state $s$. We can recursively calculate an admissible heuristic estimate $V(s)$ of the optimal value of any state $s$ in the explicit graph as follows:

$$V(s) = \begin{cases} 0 \text{ if } s \text{ is a terminal state,} \\ H(s) \text{ if } s \text{ is a nonterminal tip state,} \\ \max_{a \in A(s)} \left[ \sum_{s' \in S} Pr(s'|s,a) \left( R(s') + V(s') \right) \right] \text{ otherwise.} \end{cases} \tag{3}$$

The best partial solution graph can be determined at any time by propagating heuristic estimates from the tip states of the explicit graph to the start state. If we mark the action that maximizes the value of each state, the best partial solution graph can be determined by starting at the root of the graph and selecting the best (i.e., marked) action for each reachable state.

Table 1 outlines the AO* algorithm for finding an optimal solution graph in an acyclic AND/OR graph. It interleaves forward expansion of the best partial solution with a value update step that updates estimated state values and the best partial solution. In the simplest version of AO*, the values of the expanded state and all of its ancestor states in the explicit graph are updated. But in fact, the only ancestor states that need to be re-evaluated are those from which the expanded state can be reached by taking marked actions (*i.e.*, by choosing the best action for each state). Thus, the parenthetical remark in step 2(b)i of Table 1 indicates that a parent $s'$ of state $s$ is not added to $Z$ unless both the estimated value of state $s$ has changed and state $s$ can be reached from state $s'$ by choosing the best action for state $s'$. AO* terminates when the policy expansion step does not

1. The explicit graph $G'$ initially consists of the start state $s_0$.
2. While the best solution graph has some nonterminal tip state:
   (a) *Expand best partial solution*: Expand some nonterminal tip state $s$ of the best partial solution graph and add any new successor states to $G'$. For each new state $s'$ added to $G'$ by expanding $s$, if $s'$ is a terminal state then $V(s') := 0$; else $V(s') := H(s')$.
   (b) *Update state values and mark best actions*:
      i. Create a set $Z$ that contains the expanded state and all of its ancestors in the explicit graph along marked action arcs. (*I.e.*, only include ancestor states from which the expanded state can be reached by following the current best solution.)
      ii. Repeat the following steps until $Z$ is empty.
         A. Remove from $Z$ a state $s$ such that no descendant of $s$ in $G'$ occurs in $Z$.
         B. Set $V(s) := \max_{a \in A(s)} \sum_{s'} Pr(s'|s,a)\,(R(s') + V(s'))$ and mark the best action for $s$. (When determining the best action resolve ties arbitrarily, but give preference to the currently marked action.)
   (c) *Identify the best solution graph and all nonterminal states on its fringe*
3. Return an optimal solution graph.

Table 1: AO* algorithm.

find any nonterminal states on the fringe of the best solution graph. At this point, the best solution graph is an optimal solution.

Following the literature on AND/OR graph search, we have so far referred to the solution found by AO* as a solution graph. But in the following, when AO* is used to solve an MDP, we sometimes follow the literature on MDPs in referring to a solution as a *policy*. We also sometimes refer to it as a *policy graph*, to indicate that a policy is represented in the form of a graph.

### 3.2 Hybrid-State AO*

We now consider how to generalize AO* to solve a bounded-horizon hybrid-state MDP. The challenge we face in applying AO* to this problem is the challenge of performing state-space search in a hybrid state space.

The solution we adopt is to search in an aggregate state space that is represented by an AND/OR graph in which there is a node for each distinct value of the discrete component of the state. In other words, each node of the AND/OR graph represents a region of the continuous state space in which the discrete value is the same. Given this partition of the continuous state space, we use AND/OR graph search techniques to solve the MDP for those parts of the state space that are reachable from the start state under the best policy.

However, AND/OR graph search techniques must be modified in important ways to allow search in a hybrid state space that is represented in this way. In particular, there is no longer a correspondence between the nodes of the AND/OR graph and individual states. Each node now corresponds to a continuous region of the state space, and different actions may be optimal for different hybrid states associated with the same search node. In the case of rover planning, for example, the best action is likely to depend on how much energy or time is remaining, and energy and time are continuous state variables.

To address this problem and still find an optimal solution, we attach to each search node a set of functions (of the continuous variables) that make it possible to associate different values, heuristics, and actions with different hybrid states that map to the same search node. As before, the *explicit* *search graph* consists of all nodes and edges of the AND/OR graph that have been generated so far, and describes all the states that have been considered so far by the search algorithm. The difference is that we use a more complex state representation in which a set of continuous functions allows representation and reasoning about the continuous part of the state space associated with a search node.

We begin by describing this more complex node data structure, and then we describe the HAO* algorithm.

### 3.2.1 Data Structures

Each node $n$ of the explicit AND/OR graph $G'$ consists of the following:

- The value of the discrete state variable.
- Pointers to its parents and children in the explicit graph and the policy graph.
- $Open_n(\cdot) \rightarrow \{0,1\}$: the "Open list". For each $\mathbf{x} \in \mathbf{X}$, $Open_n(\mathbf{x})$ indicates whether $(n, \mathbf{x})$ is on the frontier of the explicit graph, i.e., generated but not yet expanded.
- $Closed_n(\cdot) \rightarrow \{0,1\}$: the "Closed list". For each $\mathbf{x} \in \mathbf{X}$, $Closed_n(\mathbf{x})$ indicates whether $(n, \mathbf{x})$ is in the interior of the explicit graph, i.e., already expanded.

  Note that, for all $(n, \mathbf{x})$, $Open_n(\mathbf{x}) \cap Closed_n(\mathbf{x}) = \emptyset$. (A state cannot be both open and closed.) There can be parts of the continuous state space associated with a node that are neither open nor closed. Until the explicit graph contains a trajectory from the start state to a particular hybrid state, that hybrid state is not considered generated, even if the search node to which it corresponds has been generated; such states are neither open nor closed. In addition, only non-terminal states can be open or closed. Note that we do not refer to open or closed nodes; instead, we refer to the hybrid states associated with nodes as being open or closed.
- $H_n(\cdot)$: the heuristic function. For each $\mathbf{x} \in \mathbf{X}$, $H_n(\mathbf{x})$ is a heuristic estimate of the optimal expected cumulative reward from state $(n, \mathbf{x})$.
- $V_n(\cdot)$: the value function. For any open state $(n, \mathbf{x})$, $V_n(\mathbf{x}) = H_n(\mathbf{x})$. For any closed state $(n, \mathbf{x})$, $V_n(\mathbf{x})$ is obtained by backing up the values of its successor states, as in Equation (4).
- $\pi_n(\cdot) \rightarrow A$: the policy. Note that it is defined for closed states only.
- $Reachable_n(\cdot) \rightarrow \{0,1\}$: For each $\mathbf{x} \in \mathbf{X}$, $Reachable_n(\mathbf{x})$ indicates whether $(n, \mathbf{x})$ is reachable by executing the current best policy beginning from the start state $(n_0, \mathbf{x}_0)$.

We assume that these various continuous functions, which represent information about the hybrid states associated with a search node, partition the state space associated with a node into a discrete number of *regions*, and associate a distinct value or action with each region. Given such a partitioning, the HAO* algorithm expands and evaluates these regions of the hybrid state space, instead of individual hybrid states. The finiteness of the partition is important in order to ensure that the search frontier can be extended by a finite number of expansions, and to ensure that HAO* can terminate after a finite number of steps. In our implementation of HAO*, described in Section 4, we use the piecewise-constant partitioning of a continuous state space proposed by Feng et al. (2004). However, any method of discrete partitioning could be used, provided that the condition above holds; for example, Li and Littman (2005) describe an alternative method of partitioning. Note that two forms of state-space partitioning are used in our algorithm. First, the hybrid state space is partitioned into a finite number of regions, one for each discrete state, where each of these regions corresponds to a node of the AND/OR graph. Second, the continuous state space associated with a particular node is further partitioned into smaller regions based on a piecewise-constant representation of a continuous function, such as the one used by Feng et al. (2004).

In addition to this more complex representation of the nodes of an AND/OR graph, our algorithm requires a more complex definition of the the best (partial) solution. In standard AO*, the one-to-one correspondence between nodes and individual states means that a solution or policy can be represented entirely by a graph, called the (partial) solution graph, in which a single action is associated with each node. In the HAO* algorithm, a continuum of states is associated with each node, and different actions may be optimal for different regions of the state space associated with a particular node. For the HAO* algorithm, a (partial) solution graph is a sub-graph of the explicit graph that is defined as follows:

- the start node belongs to a solution graph;
- for every non-tip node in a solution graph, one *or more* outgoing $k$-connectors are part of the solution graph, one for each action that is optimal for some hybrid state associated with the node, and each of their successor nodes also belongs to the solution graph;
- every directed path in the solution graph terminates at a tip node of the explicit graph.

The key difference in this definition is that there may be more than one optimal action associated with a node, since different actions may be optimal for different hybrid states associated with the node. A policy is represented not only by a solution graph, but by the continuous functions $\pi_n(.)$ and $Reachable_n(.)$. In particular, a (partial) policy $\pi$ specifies an action for each reachable region of the continuous state space. The best (partial) policy is the one that satisfies the following optimality equation:

$$\begin{aligned} V_n(\mathbf{x}) &= 0 \text{ when } (n, \mathbf{x}) \text{ is a terminal state,} \\ V_n(\mathbf{x}) &= H_n(\mathbf{x}) \text{ when } (n, \mathbf{x}) \text{ is a nonterminal open state,} \\ V_n(\mathbf{x}) &= \max_{a \in A_n(\mathbf{x})} \left[ \sum_{n' \in N} \Pr(n' \mid n, \mathbf{x}, a) \int_{\mathbf{x}'} \Pr(\mathbf{x}' \mid n, \mathbf{x}, a, n') \left( R_{n'}(\mathbf{x}') + V_{n'}(\mathbf{x}') \right) d\mathbf{x}' \right]. \end{aligned} \tag{4}$$

Note that this optimality equation is only satisfied for regions of the state space that are reachable from the start state, $(n_0, \mathbf{x}_0)$ by following an optimal policy.

#### 3.2.2 Algorithm

Table 2 gives a high-level summary of the HAO* algorithm. In outline, it is the same as the AO* algorithm, and consists of iteration of the same three steps; solution (or policy) expansion, use of dynamic programming to update the current value function and policy, and analysis of reachability to identify the frontier of the solution that is eligible for expansion. In detail, it is modified in several important ways to allow search of a hybrid state space. In the following, we discuss the modifications to each of these three steps.

**Policy expansion** All nodes of the current solution graph are identified and one or more open regions associated with these nodes are selected for expansion. That is, one or more regions of the hybrid state space in the intersection of Open and Reachable is chosen for expansion. All actions applicable to the states in these open regions are simulated, and the results of these actions are added to the explicit graph. In some cases, this means adding a new node to the AND/OR graph. In other cases, it simply involves marking one or more regions of the continuous state space associated with an existing node as open. More specifically, when an action leads to a new node, this node is added to the explicit graph, and all states corresponding to this node that are reachable from the expanded region(s) after the action under consideration are marked as open. When an action leads to an

1. The explicit graph $G'$ initially consists of the start node and corresponding start state $(n_0, \mathbf{x}_0)$, marked as open and reachable.
2. While $Reachable_n(\mathbf{x}) \cap Open_n(\mathbf{x})$ is non-empty for some $(n, \mathbf{x})$:
   (a) *Expand best partial solution*: Expand one or more region(s) of open states on the frontier of the explicit state space that is reachable by following the best partial policy. Add new successor states to $G'$. In some cases, this requires adding a new node to the AND/OR graph. In other cases, it simply involves marking one or more regions of the continuous state space associated with an existing node as open. States in the expanded region(s) are marked as closed.
   (b) *Update state values and mark best actions*:
      i. Create a set $Z$ that contains the node(s) associated with the just expanded regions of states and all ancestor nodes in the explicit graph along marked action arcs.
      ii. Decompose the part of the explicit AND/OR graph that consists of nodes in $Z$ into strongly connected components.
      iii. Repeat the following steps until $Z$ is empty.
         A. Remove from $Z$ a set of nodes such that (1) they all belong to the same connected component, and (2) no descendant of these nodes occurs in $Z$.
         B. For every node $n$ in this connected component and for all states $(n, \mathbf{x})$ in any expanded region of node $n$, set

$$V_n(\mathbf{x}) := \max_{a \in A_n(\mathbf{x})} \left[ \sum_{n' \in N} \Pr(n' \mid n, \mathbf{x}, a) \int_{\mathbf{x}'} \Pr(\mathbf{x}' \mid n, \mathbf{x}, a, n') \left( R_{n'}(\mathbf{x}') + V_{n'}(\mathbf{x}') \right) d\mathbf{x}' \right],$$

         and mark the best action. (When determining the best action resolve ties arbitrarily, but give preference to the currently marked action.) Repeat until there is no longer a change of value for any of these nodes.
   (c) *Identify the best solution graph and all nonterminal states on its frontier.* This step updates $Reachable_n(\mathbf{x})$.
3. Return an optimal policy.

Table 2: HAO* algorithm.

existing node, any region(s) of Markov states in this node that is both reachable from the expanded region(s) and not marked as closed, is marked open. Expanded regions of the state space are marked as closed. Thus, different regions associated with the same node can be opened and expanded at different times. This process is illustrated in Figure 2. In this figure, nodes corresponding to a distinct value for the discrete state are represented as rectangles, and circular connectors represent actions. For each node, we see how many distinct continuous regions exist. For each such region we see whether it is closed ("C") or open ("O"), and whether it is reachable from the initial state ("R") when executing the current best policy ("OPT"). For instance, in Figure 2(a), node At(Start) has a single region marked closed and reachable, and node Lost has two regions: the smallest, open and reachable, and the largest, closed and unreachable.

**Dynamic programming** As in standard AO*, the value of any newly-expanded node $n$ must be updated by computing a Bellman backup based on the value functions of the children of $n$ in the explicit graph. For each expanded region of the state space associated with node $n$, each action is evaluated, the best action is selected, and the corresponding continuous value function is associated with the region. The continuous-state value function is computed by evaluating the continuous integral in Equation (4). We can use any method for computing this integral. In our implementation, we use the dynamic programming algorithm of Feng et al. (2004). As reviewed in Section 2.4, they show that the continuous integral over $\mathbf{x}'$ can be computed exactly, as long as the transition and reward functions satisfy certain conditions. Note that, with some hybrid-state dynamic programming techniques such as Feng et al. (2004), dynamic programming backups may increase the number of pieces of the value function attached to the updated regions (Figure 3(a)).

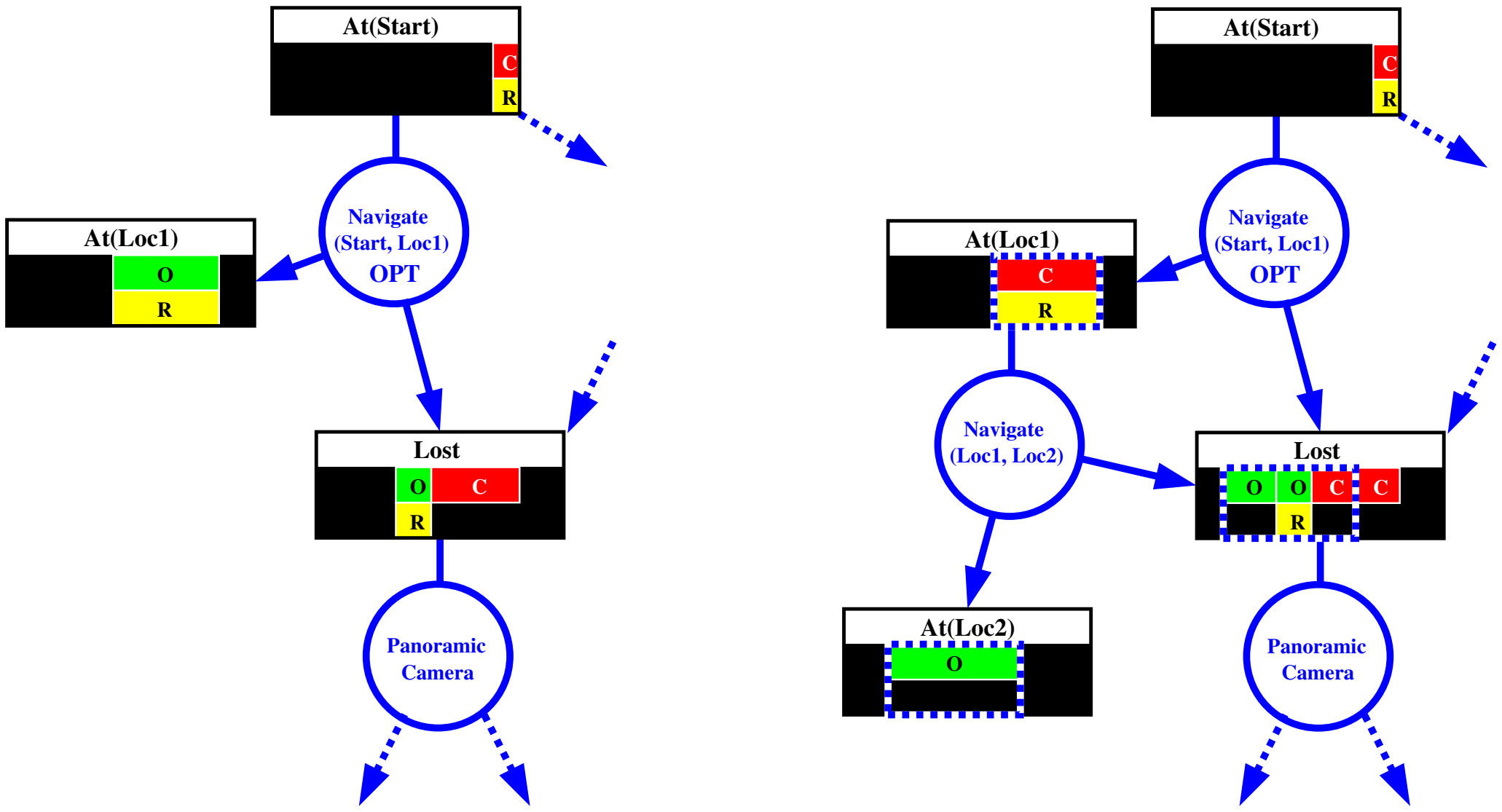

(a) Before expansion (b) After expansion

Figure 2: Expanding a region of the state space. (a) Before expansion: The nodes At(Start), At(Loc1) and Lost have been previously created. The unique region in At(Loc1) is the next region to be expanded. (b) After expansion: The action Navigate(Loc1, Loc2) that can be applied in the expanded region has been added to the graph. This action can lead either to the preexisting node Lost, or to the new node At(Loc2). The expanded region (in At(Loc1)), as well as the continuous regions reachable from there (in Lost and At(Loc2)), are highlighted in a dotted framed. Following expansion, the expanded region is closed. Discrete state At(Loc2) has been added to the graph and all its reachable regions are open. Additionally, new open regions have been added to node Lost.

Once the expanded regions of the continuous state space associated with a node $n$ are re-evaluated, the new values must be propagated backward in the explicit graph. The backward propagation stops at nodes where the value function is not modified, or at the root node. The standard AO* algorithm, summarized in Figure 1, assumes that the AND/OR graph in which it searches is acyclic. There are extensions of AO* for searching in AND/OR graphs that contain cycles. One line of research is concerned with how to find acyclic solutions in AND/OR graphs that contain cycles (Jimenez & Torras, 2000). Another generalization of AO*, called LAO*, allows solutions to contain cycles or “loops” in order to specify policies for infinite-horizon MDPs (Hansen & Zilberstein, 2001).

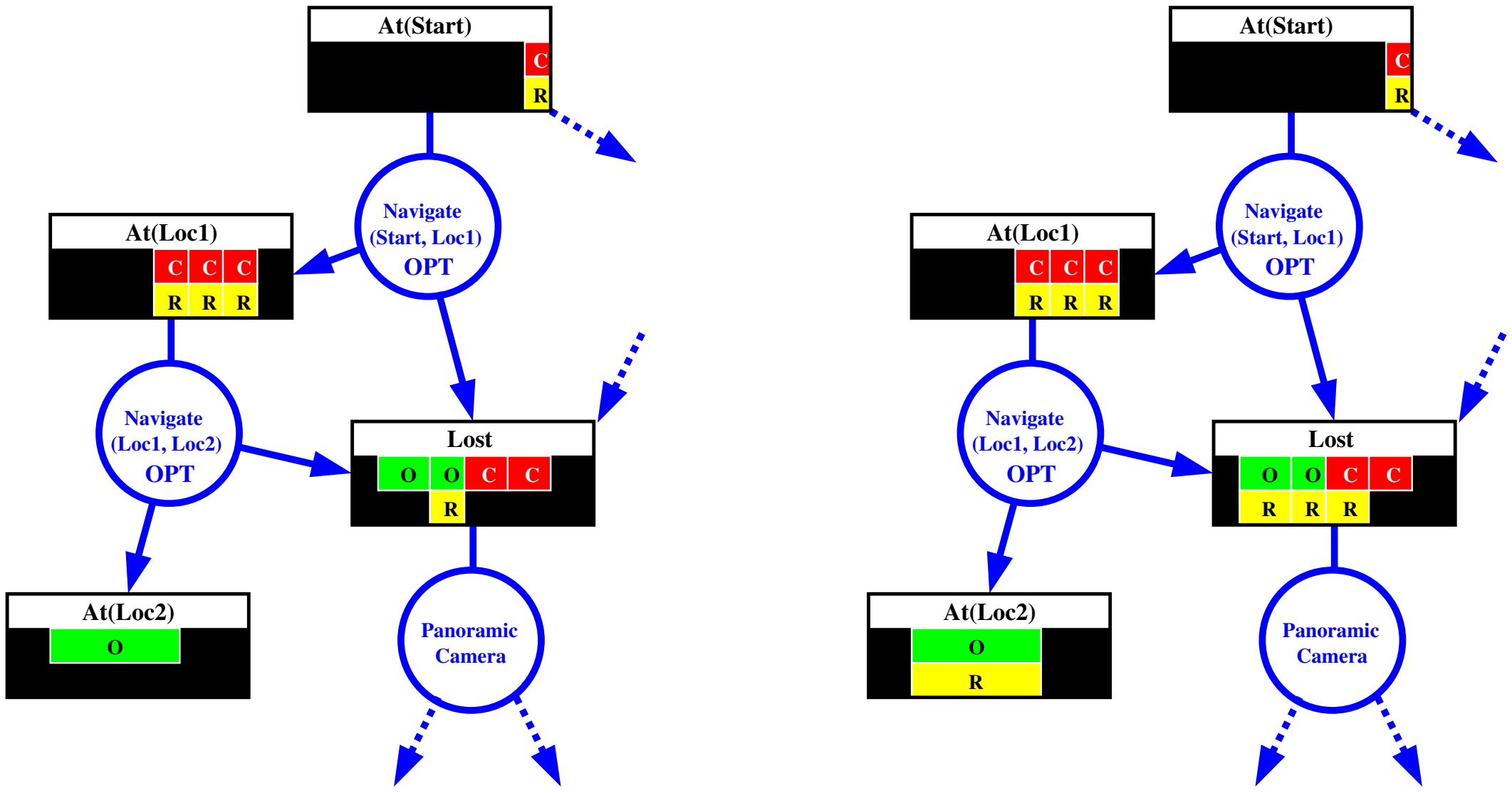

(a) Dynamic programming

(b) Reachability analysis

Figure 3: Dynamic programming and reachability analysis (Figure 2 continued). (a) Dynamic programming: The optimal policy has been reevaluated and Navigate(Loc1, Loc2) appears optimal in some continuous states of At(Loc2). Node At(Loc1) is represented with a finer partition of the continuous state space to illustrate the fact that the backup increased the number of pieces of the value function associated with the expanded region. (b) Reachability analysis: The newly created region of At(Loc2) becomes reachable, as well as the regions of Lost that can be reached through Navigate(Loc1, Loc2).

Given our assumption that every action has positive resource consumption, there can be no loops in the state space of our problem because the resources available decrease at each step. But surprisingly, there can be loops in the AND/OR graph. This is possible because the AND/OR graph represents a projection of the state space onto a smaller space that consists of only the discrete component of the state. For example, it is possible for the rover to return to the same site it has visited before. The rover is not actually in the same state, since it has fewer resources available. But the AND/OR graph represents a projection of the state space that does not include the continuous aspects of the state, such as resources, and this means the rover can visit a state that projects to the same node of the AND/OR graph as a state it visited earlier, as shown in Figure 4. As a result, there can be loops in the AND/OR graph, and even loops in the part of the AND/OR graph that corresponds to a solution. But in a sense, these are "phantom loops" that can only appear in the projected state space, and not in the real state space.

Nevertheless we must modify the dynamic programming (DP) algorithm to deal with these loops. Because there are no loops in the real state space, we know that the exact value function can be updated by a finite number of backups performed in the correct order, with one backup performed for any state that can be visited along a path from the start state to the expanded node(s). But because multiple states can map to the same AND/OR graph node, the continuous region of the state space associated with a particular node may need to be evaluated more than once. To identify the AND/OR graph nodes that need to be evaluated more than once, we use the following two-step algorithm.

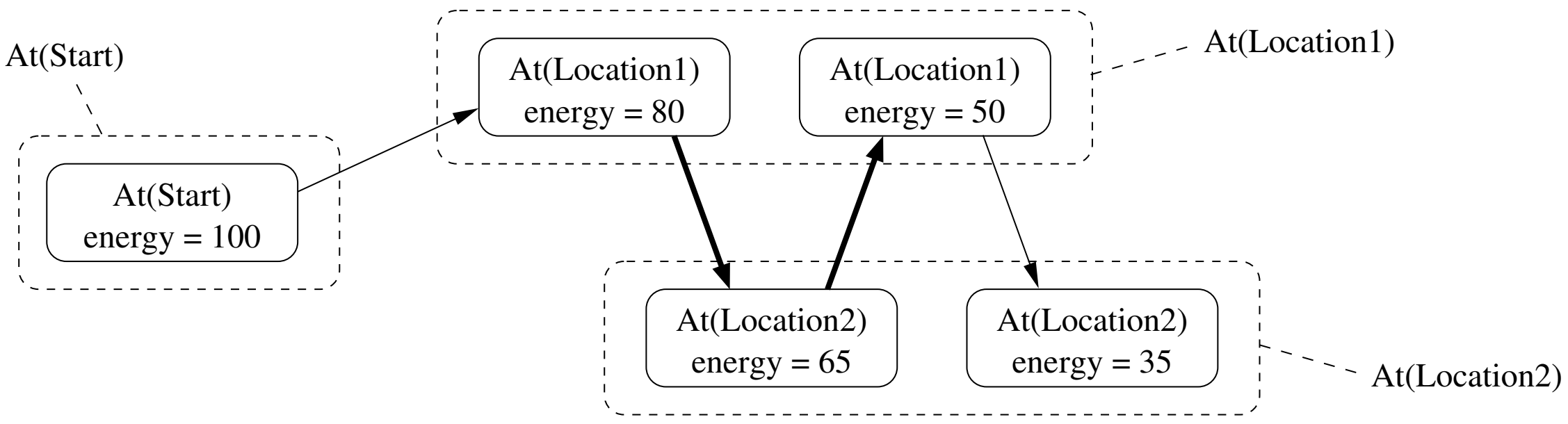

Figure 4: Phantom loops in HAO*: solid boxes represent Markov states. Dashed boxes represent search nodes, that is, the projection of Markov states on the discrete components. Arrows represent possible state transition. Bold arrows show an instance of phantom loop in the search space.

First, we consider the part of the AND/OR graph that consists of ancestor nodes of the just expanded node(s). This is the set $Z$ of nodes identified at the beginning of the DP step. We decompose this part of the graph into strongly connected components. The graph of strongly connected components is acyclic and can be used to prescribe the order of backups in almost the same way as in the standard AO* algorithm. In particular, the nodes in a particular component are not backed up until all nodes in its descendant components have been backed up. Note that in the case of an acyclic graph, every strongly connected component has a single node. It is only possible for a connected component to have more than one node if there are loops in the AND/OR graph.

If there are loops in the AND/OR graph, the primary change in the DP step of the algorithm occurs when it is time to perform backups on the nodes in a connected component with more than one node. In this case, all nodes in the connected component are evaluated. Then, they are repeatedly re-evaluated until the value functions of these nodes converge, that is, until there is no change in the values of any of the nodes. Because there are no loops in the real state space, convergence is guaranteed to occur after a finite number of steps. Typically, it occurs after a very small number of steps. An advantage of decomposing the AND/OR graph into connected components is that it identifies loops and localizes their effect to a small number of nodes. In experiments in our test domain, most nodes of the graph need to be evaluated just once during the DP step, and only a small number of nodes (and often none) need to be evaluated more than once.

Note that decomposition of the nodes in $Z$ into connected components is a method for improving the efficiency of the dynamic programming step, and is not required for its correctness. The alternative of repeatedly updating all nodes in $Z$ until all their values converge is also correct, although it is likely to result in many useless updates of already converged nodes.

**Analysis of reachability** Change in the value function can lead to change in the optimal policy, and, thus, to change in which states are visited by the best policy. This, in turn, can affect which open regions of the state space are eligible to be expanded. In this final step, HAO* identifies the best (partial) policy and recomputes $\mathit{Reachable}_n$ for all nodes and states in the explicit graph, as follows (see Figure 3(b)). For each node $n$ in the best (partial) solution graph, consider each of its parents $n'$ in the solution graph, and all the actions $a$ that can lead from one of the parents to $n$. Then $\mathit{Reachable}_n(\mathbf{x})$ is the support of $P_n(\mathbf{x})$, where

$$P_n(\mathbf{x}) = \sum_{(n',a)\in\Omega_n} \int_{\mathbf{X}} \mathit{Reachable}_{n'}(\mathbf{x}') \Pr(n \mid n', \mathbf{x}', a) \Pr(\mathbf{x} \mid n', \mathbf{x}', a, n) d\mathbf{x}' \,, \tag{5}$$

that is, $Reachable_n(\mathbf{x}) = \{\mathbf{x} \in \mathbf{X} : P_n(\mathbf{x}) > 0\}$. In Equation (5), $\Omega_n$ is the set of pairs $(n', a)$ where $a$ is the best action in $n'$ for some reachable resource level:

$$\Omega_n = \{(n', a) \in N \times A : \exists \mathbf{x} \in \mathbf{X}, \quad P_{n'}(\mathbf{x}) > 0, \ \pi_{n'}(\mathbf{x}) = a, \ \Pr(n \mid n', \mathbf{x}, a) > 0\} \ .$$

It is clear that we can restrict our attention to state-action pairs in $\Omega_n$, only.

By performing this reachability analysis, HAO* identifies the frontier of the state space that is eligible for expansion. HAO* terminates when this frontier is empty, that is, when it does not find any hybrid states in the intersection of *Reachable* and *Open*.

### 3.3 Convergence and Error Bounds

We next consider some of the theoretical properties of HAO*. First, under reasonable assumptions, we prove that HAO* converges to an optimal policy after a finite number of steps. Then we discuss how to use HAO* to find sub-optimal policies with error bounds.

The proof of convergence after a finite number of steps depends, among other things, on the assumption that a hybrid-state MDP has a finite branching factor. In our implementation, this means that for any region of the state space that can be represented by a hyper-rectangle, the set of successor regions after an action can be represented by a finite set of hyper-rectangles. From this assumption and the assumption that the number of actions is finite, it follows that for every assignment $n$ to the discrete variables, the set

$$\{\mathbf{x} | (n, \mathbf{x}) \text{is reachable from the initial state using some fixed sequence of actions}\}$$

is the union of a finite number of open or closed hyper-rectangles. This assumption can be viewed as a generalization of the assumption of a finite branching factor in a discrete AND/OR graph upon which the finite convergence proof of AO* depends.

**Theorem 1** *If the heuristic functions $H_n$ are admissible (optimistic), all actions have positive resource consumptions, both continuous backups and action application are computable exactly in finite time, and the branching factor is finite, then:*

1. *At each step of HAO*, $V_n(\mathbf{x})$ is an upper-bound on the optimal expected return in $(n, \mathbf{x})$, for all $(n, \mathbf{x})$ expanded by HAO*;*
2. *HAO* terminates after a finite number of steps;*
3. *After termination, $V_n(\mathbf{x})$ is equal to the optimal expected return in $(n, \mathbf{x})$, for all $(n, \mathbf{x})$ reachable under an optimal policy, i.e., $Reachable_n(\mathbf{x}) > 0$.*

*Proof*: (1) The proof is by induction. Every state $(n, \mathbf{x})$ is assigned an initial heuristic estimate, and $V_n(\mathbf{x}) = H_n(\mathbf{x}) \geq V_n^*(\mathbf{x})$ by the admissibility of the heuristic evaluation function. We make the inductive hypothesis that at some point in the algorithm, $V_n(\mathbf{x}) \geq V_n^*(\mathbf{x})$ for every state $(n, \mathbf{x})$. If a backup is performed for any state $(n, \mathbf{x})$,

$$\begin{aligned}
V_n(\mathbf{x}) &= \max_{a \in A_n(\mathbf{x})} \left[ \sum_{n' \in N} \Pr(n' \mid n, \mathbf{x}, a) \int_{\mathbf{x}'} \Pr(\mathbf{x}' \mid n, \mathbf{x}, a, n') \left( R_{n'}(\mathbf{x}') + V_{n'}(\mathbf{x}') \right) d\mathbf{x}' \right] \\
&\geq \max_{a \in A_n(\mathbf{x})} \left[ \sum_{n' \in N} \Pr(n' \mid n, \mathbf{x}, a) \int_{\mathbf{x}'} \Pr(\mathbf{x}' \mid n, \mathbf{x}, a, n') \left( R_{n'}(\mathbf{x}') + V^*_{n'}(\mathbf{x}') \right) d\mathbf{x}' \right] \\
&= V_n^*(\mathbf{x}) \,,
\end{aligned}$$

where the last equality restates the Bellman optimality equation.

(2) Because each action has positive, bounded from below, resource consumption, and resources are finite and non-replenishable, the complete implicit AND/OR graph must be finite. For the same reason, this graph can be turned into a finite graph without loops: Along any directed loop in this graph, the amount of maximal available resources must decrease by some $\epsilon$ which is a positive lower-bound on the amount of resources consumed by an action. Each node in this graph may be expanded a number of times that is bounded by the number of its ancestor. (Each time a new ancestor is discovered, it may lead to an update in the set of reachable regions for this node.) Moreover, finite branching factor implies that the number of regions considered within each node is bounded (because there are finite ways of reaching this node, each of which contributes a finite number of hyper-rectangles). Thus, overall, the number of regions considered is finite, and the processing required for each region expansion is finite (because action application and backups are computed in finite time). This leads to the desired conclusion.

(3) The search algorithm terminates when the policy for the start state $(n_0, \mathbf{x}_0)$ is complete, that is, when it does not lead to any unexpanded states. For every state $(n, \mathbf{x})$ that is reachable by following this policy, it is contradictory to suppose $V_n(\mathbf{x}) > V_n^*(\mathbf{x})$ since that implies a complete policy that is better than optimal. By the Bellman optimality equation of Equation (1), we know that $V_n(\mathbf{x}) \geq V_n^*(\mathbf{x})$ for every state in this complete policy. Therefore, $V_n(\mathbf{x}) = V_n^*(\mathbf{x})$. □

HAO* not only converges to an optimal solution, stopping the algorithm early allows a flexible trade-off between solution quality and computation time. If we assume that, in each state, there is a *done* action that terminates execution with zero reward (in a rover problem, we would then start a safe sequence), then we can evaluate the current policy at each step of the algorithm by assuming that execution ends each time we reach a leaf of the policy graph. Under this assumption, the error of the current policy at each step of the algorithm can be bounded. We show this by using a decomposition of the value function described by Chakrabarti et al.(1988) and Hansen and Zilberstein (2001). We note that at any point in the algorithm, the value function can be decomposed into two parts, $g_n(\mathbf{x})$ and $h_n(\mathbf{x})$, such that

$$\begin{aligned} g_n(\mathbf{x}) &= 0 \text{ when } (n, \mathbf{x}) \text{ is an open state, on the fringe of the greedy policy; otherwise,} \\ g_n(\mathbf{x}) &= \sum_{n' \in N} \Pr(n' \mid n, \mathbf{x}, a^*) \int_{\mathbf{x}'} \Pr(\mathbf{x}' \mid n, \mathbf{x}, a^*, n') \left(R_n(\mathbf{x}) + g_{n'}(\mathbf{x}')\right) d\mathbf{x}' , \end{aligned} \tag{6}$$

and

$$\begin{aligned} h_n(\mathbf{x}) &= H_n(\mathbf{x}) \text{ when } (n, \mathbf{x}) \text{ is an open state, on the fringe of the greedy policy; otherwise,} \\ h_n(\mathbf{x}) &= \sum_{n' \in N} \Pr(n' \mid n, \mathbf{x}, a^*) \int_{\mathbf{x}'} \Pr(\mathbf{x}' \mid n, \mathbf{x}, a^*, n')\; h_{n'}(\mathbf{x}') d\mathbf{x}' , \end{aligned} \tag{7}$$

where $a^*$ is the action that maximizes the right-hand side of Equation (4). Note that $V_n(\mathbf{x}) = g_n(\mathbf{x}) + h_n(\mathbf{x})$. We use this decomposition of the value function to bound the error of the best policy found so far, as follows.

**Theorem 2** *At each step of the HAO* algorithm, the error of the current best policy is bounded by* $h_{n_0}(\mathbf{x_0})$.

*Proof*: For any state $(n, \mathbf{x})$ in the explicit search space, a lower bound on its optimal value is given by $g_n(\mathbf{x})$, which is the value that can be achieved by the current policy when the *done* action is executed at all fringe states, and an upper bound is given by $V_n(\mathbf{x}) = g_n(\mathbf{x}) + h_n(\mathbf{x})$, as established in Theorem 1. It follows that $h_{n_0}(\mathbf{x_0})$ bounds the difference between the optimal value and the current admissible value of any state $(n, \mathbf{x})$, including the initial state $(n_0, \mathbf{x}_))$.□

Note that the error bound for the initial state is $h_{n_0}(\mathbf{x_0}) = H_{n_0}(\mathbf{x_0})$ at the start of the algorithm; it decreases with the progress of the algorithm; and $h_{n_0}(\mathbf{x_0}) = 0$ when HAO* converges to an optimal solution.

### 3.4 Heuristic Function

The heuristic function $H_n$ focuses the search on reachable states that are most likely to be useful. The more informative the heuristic, the more scalable the search algorithm. In our implementation of HAO* for the rover planning problem, which is described in detail in the next section, we used the simple admissible heuristic function which assigns to each node the sum of all rewards associated with goals that have not been achieved so far. Note that this heuristic function only depends on the discrete component of the state, and not on the continuous variables; that is, the function $H_n(\mathbf{x})$ is constant over all values of $\mathbf{x}$. It is obvious that this heuristic is admissible, since it represents the maximum additional reward that could be achieved by continuing plan execution. Although it is not obvious that a heuristic this simple could be useful, the experimental results we present in Section 4 show that it is. We considered an additional, more informed heuristic function that solved a relaxed, suitably discretized, version of the planning problem. However, taking into account the time required to compute this heuristic estimate, the simpler heuristic performed better.

### 3.5 Expansion Policy

HAO* works correctly and converges to an optimal solution no matter which continuous region(s) of which node(s) are expanded in each iteration (step 2.a). But the quality of the solution may improve more quickly by using some "heuristics" to choose which region(s) on the fringe to expand next.

One simple strategy is to select a node and expand all continuous regions of this node that are open and reachable. In a preliminary implementation, we expanded (the open regions of) the node that is most likely to be reached using the current policy. Changes in the value of these states will have the greatest effect on the value of earlier nodes. Implementing this strategy requires performing the additional work involved in maintaining the probability associated with each state. If such probabilities are available, one could also focus on expanding the most promising node, that is, the node where the integral of $H_n(\mathbf{x})$ times the probability over all values of $\mathbf{x}$ is the highest, as described by Mausam, Benazera, Brafman, Meuleau, and Hansen (2005).

Hansen and Zilberstein (2001) observed that, in the case of LAO*, the algorithm is more efficient if we expand several nodes in the fringe before performing dynamic programming in the explicit graph. This is because the cost of performing the update of a node largely dominates the cost of expanding a node. If we expand only one node of the fringe at each iteration, we might have to perform more DP backups than if we expand several nodes with common ancestors before proceeding to DP. In the limit, we might want to expand all nodes of the fringe at each algorithm iteration. Indeed, this variant of LAO* proved the most efficient (Hansen & Zilberstein, 2001).

In the case of LAO*, updates are expensive because of the loops in the implicit graph. In HAO*, the update of a region induces a call to the hybrid dynamic programming module for each open region of the node. Therefore, the same technique is likely to produce the same benefit.

Pursuing this idea, we allowed our algorithm to expand all nodes in the fringe and all their descendants up to a fixed depth at each iteration. We defined a parameter, called the *expansion horizon* and denoted $k$, to represent, loosely speaking, the number of times the whole fringe is expanded at each iteration. When $k = 1$, HAO* expands all open and reachable regions of all nodes in the fringe before recomputing the optimal policy. When $k = 2$, it expands all regions in the fringe and all their children before updating the policy. At $k = 3$ it also consider the grand-children of regions in the fringe, and so on. When $k$ tends to infinity, the algorithm essentially performs an exhaustive search: it first expands the graph of all reachable nodes, then performs one pass of (hybrid) dynamic programming in this graph to determine the optimal policy. By balancing node expansion and update, the expansion horizon allows tuning the algorithm behavior from an exhaustive search to a more traditional heuristic search. Our experiments showed that a value of $k$ between 5 and 10 is optimal to solve our hardest benchmark problems (see section 4).

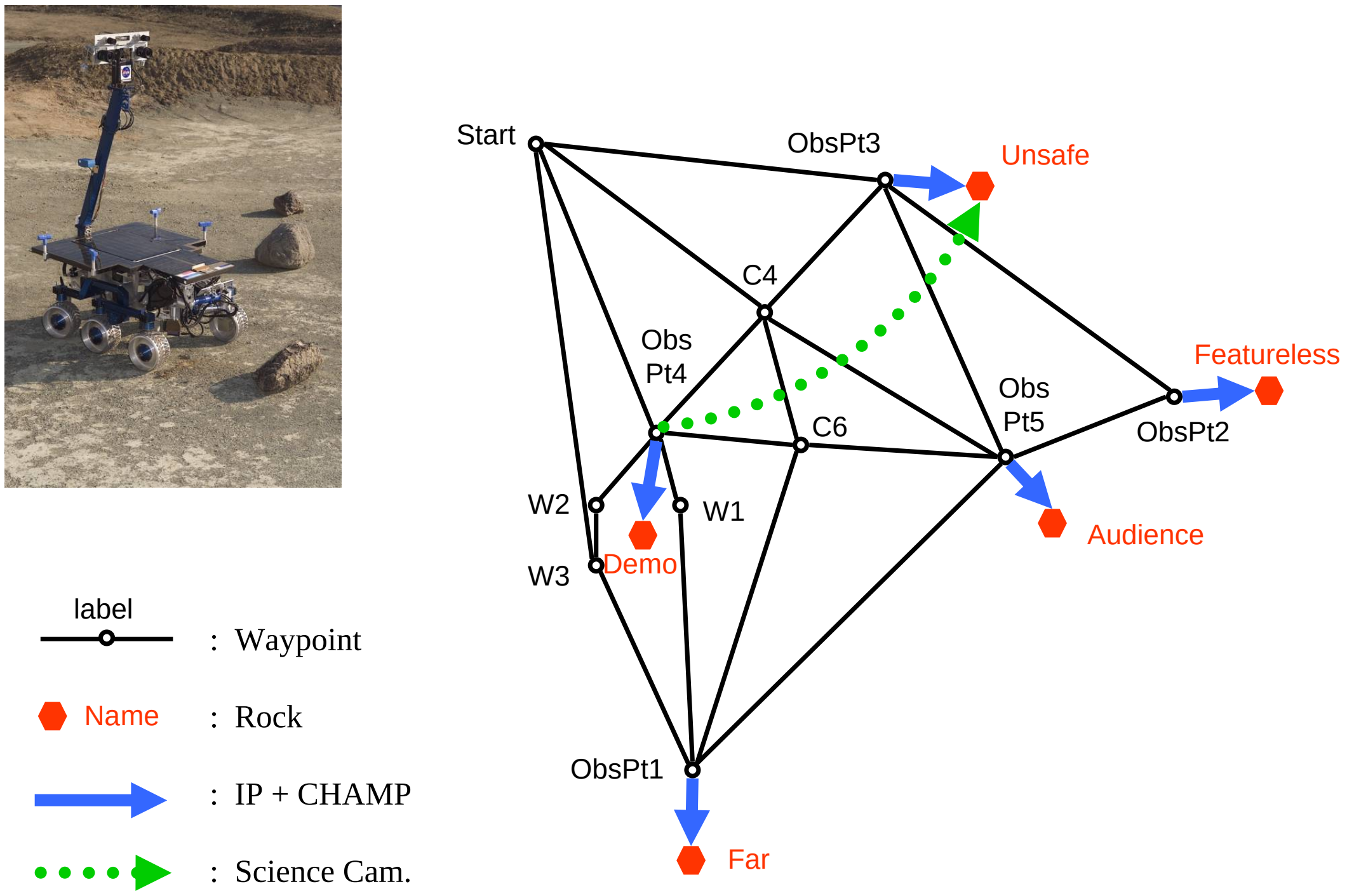

Figure 5: The K9 rover (top left) was developed at the Jet Propulsion Laboratory and NASA Ames Research Center as a prototype for the MER rovers. It is used to test advanced rover software, including automated planners of the rover's activities. Right: topological map of the 2004 IS demo problem. Arrows labeled "IP + CHAMP" represent the opportunity to deploy the arm against a rock (instrument placement) and take a picture of it with the CHAMP Camera. Arrows labeled "Science Cam" represent the opportunity to take a remote picture of a rock with the Science Camera.

### 3.6 Updating Multiple Regions

The expansion policies described above are based on expanding all open regions of one or several nodes simultaneously. They allow leveraging hybrid-state dynamic programming techniques such as those of Feng et al. (2004) and Li and Littman (2005). These techniques may compute in a single iteration piecewise constant and linear value functions that cover a large range of continuous states, possibly the whole space of possible values. In particular, they can back up in one iteration all continuous states included between given bounds.

Therefore, when several open regions of the same node are expanded at the same iteration of HAO*, we can update all of them simultaneously by backing-up a subset of continuous states that includes all these regions. For instance, one may record lower bounds and upper bounds on each continuous variable over the expanded regions, and then compute a value function that covers the hyper-rectangle between these bounds.

This modification of the algorithm does not impact convergence. As long as the value of all expanded regions is computed, the convergence proof holds. However, execution time may be adversely affected if the expanded regions are a proper subset of the region of continuous states that is

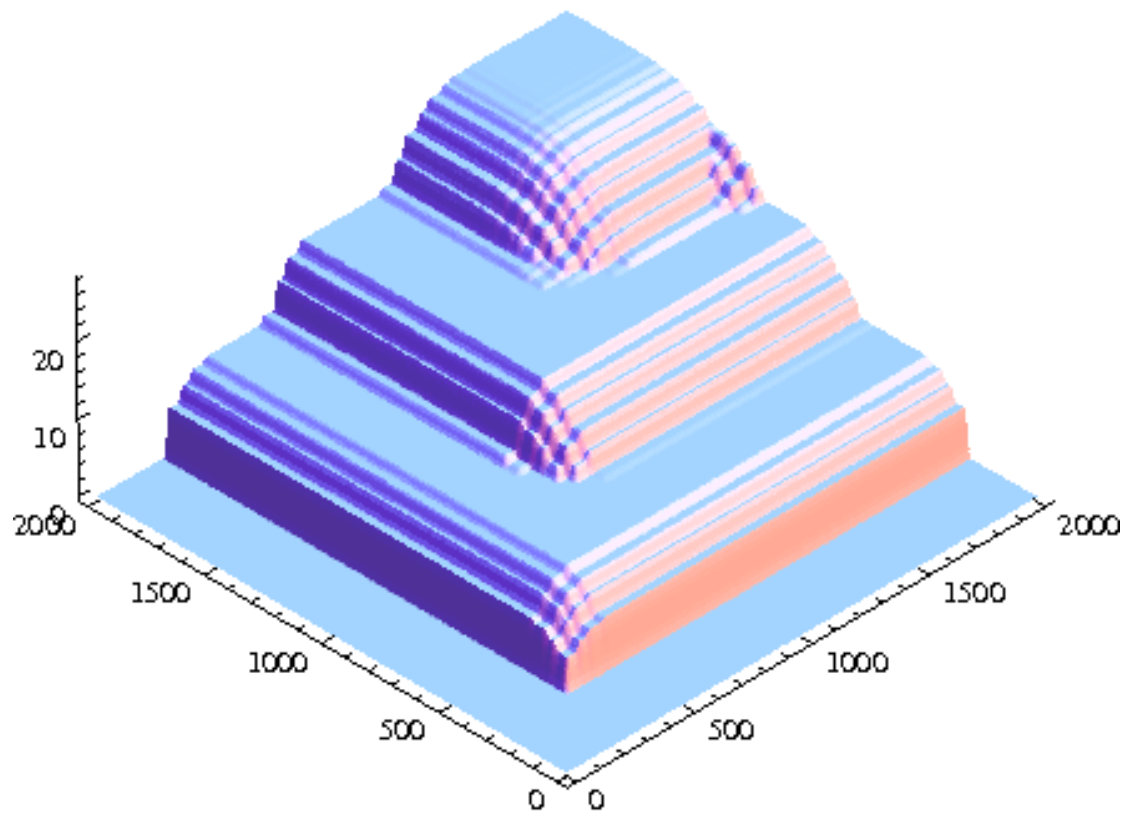

(a) Value function $V_n(.)$ for the initial node. The first plateau corresponds to analyzing R1, the second plateau to analyzing R2, and the third plateau to analyzing both R1 and R2.

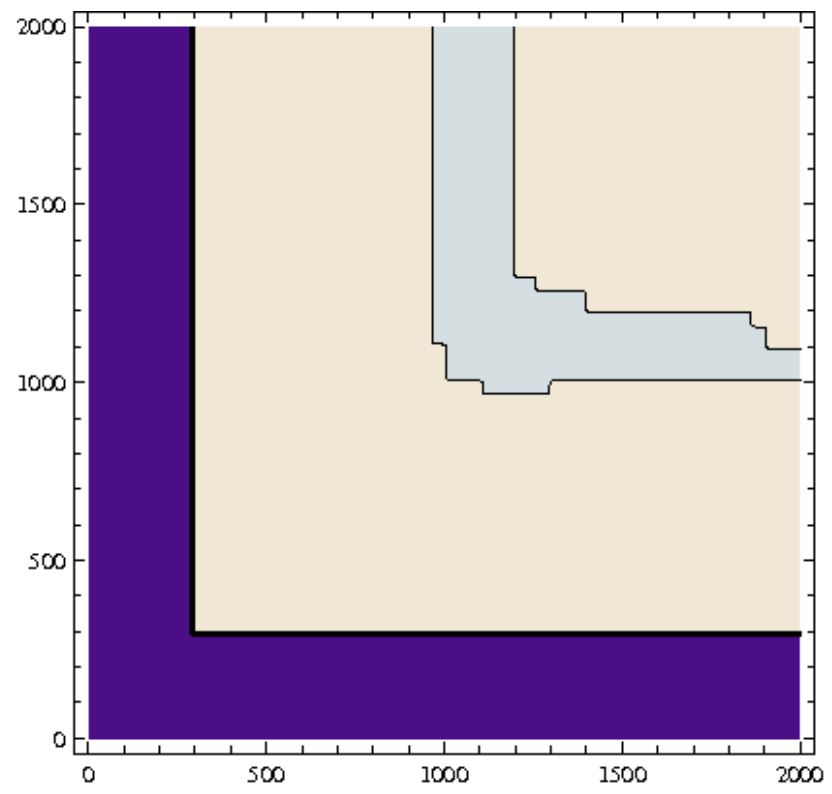

(b) The policy $\pi_n(.)$ for the starting node shows the partitions of the resource space where different actions are optimal. Dark: no action; Grey: navigation to R2; Light: analysis of R1.

Figure 6: (a) Optimal value function for the initial state of the simple rover problem over all possible values for the continuous resources (time and energy remaining). The value function is partitioned into 3476 pieces. (b) Optimal policy for the same set of states.

backed-up. In that case, the values of states that are not open or not reachable is uselessly computed, which deviates from a pure heuristic search algorithm.

However, this modification may also be beneficial because it avoids some redundant computation. Hybrid-state dynamic programming techniques manipulate pieces of value functions. Thus, if several expanded regions are included in the same piece of the value function, their value is computed only once. In practice, this benefit may outweigh the cost of evaluating useless regions. Moreover, cost is further reduced by storing the value functions associated with each node of the graph, so that computed values of irrelevant regions are saved in case these regions become eligible for expansion (i.e., open and reachable) later. Thus, this variant of HAO* fully exploits hybrid-state dynamic programming techniques.

## 4. Experimental Evaluation

In this section, we describe the performance of HAO* in solving planning problems for a simulated planetary exploration rover with two monotonic and continuous-valued resources: time and battery power. Section 4.1 uses a simple "toy" example of this problem to illustrate the basic steps of the HAO* algorithm. Section 4.2 tests the performance of the algorithm using a realistic, real-size NASA simulation of a rover and analyzes the results of the experiments. The simulation uses a model of the K9 rover (see Figure 5) developed for the Intelligent Systems (IS) demo at NASA Ames Research Center in October 2004 (Pedersen et al., 2005). This is a complex real-size model of the K9 rover that uses command names understandable by the rover's execution language, so that the plans produced by our algorithm can be directly executed by the rover. For the experiments reported in Section 4.2, we did not simplify this NASA simulation model in any way.

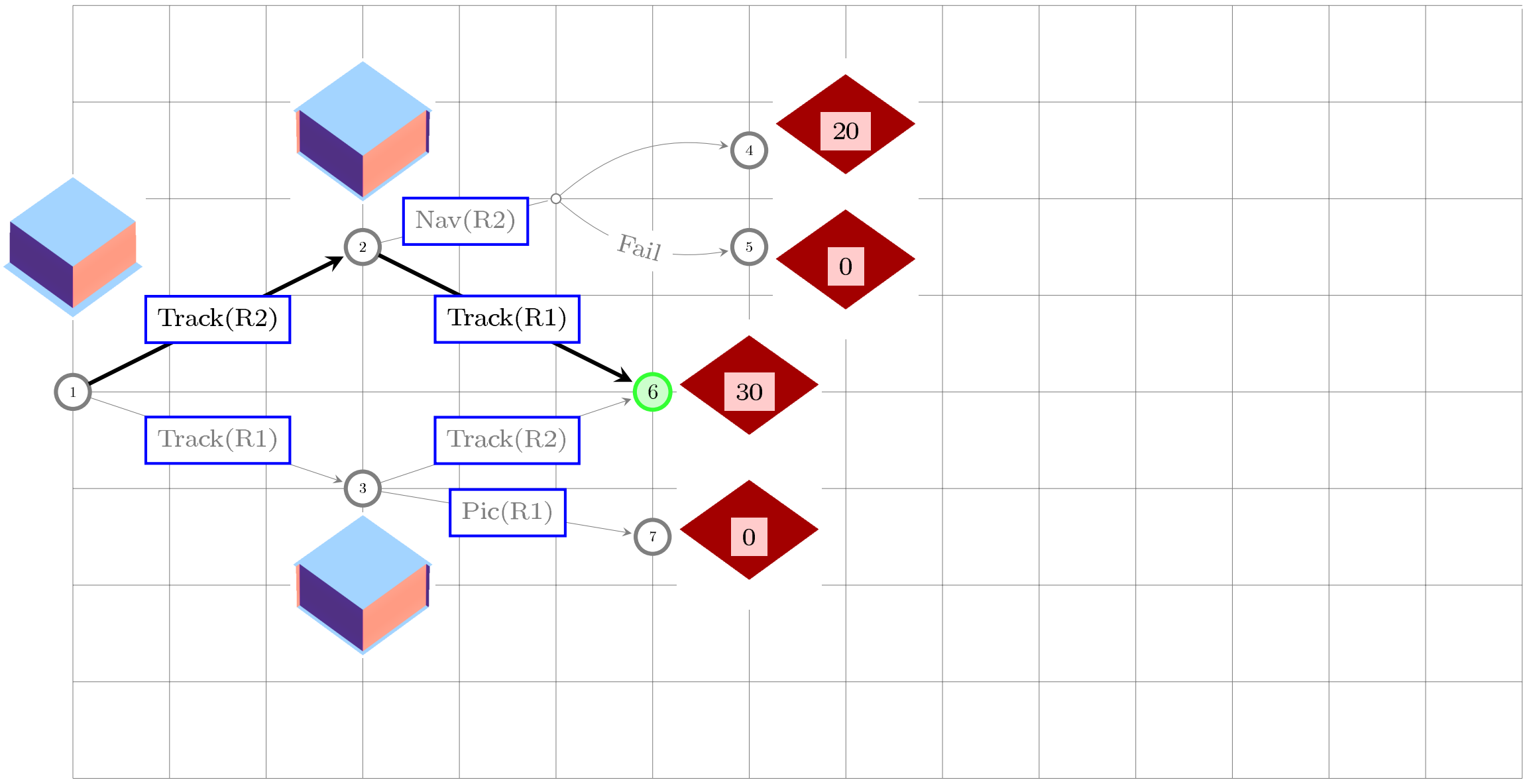

Figure 7: First iteration of HAO* on the toy problem. The explicit graph is marked by dim edges and the solution graph is marked by thick edges. Tip nodes 4, 5, 6 and 7 are shown with constant heuristic functions and expanded nodes 1, 2 and 3 are shown with backed up value functions.

In the planning problem we consider, an autonomous rover must navigate in a planar graph representing its surroundings and the authorized navigation paths, and schedule observations to be performed on different rocks situated at different locations. Only a subset of its observational goals can be achieved in a single run due to limited resources. Therefore, this is an oversubscribed planning problem. It is also a problem of planning under uncertainty since each action has uncertain positive resource consumptions and a probability of failing.

A significant amount of uncertainty in the domain comes from the tracking mechanism used by the rover. Tracking is the process by which the rover recognizes a rock based on certain features in its camera image that are associated with the rock. During mission operations, a problem instance containing a fixed set of locations, paths, and rocks is built from the last panoramic camera image sent by the rover. Each “logical rock” in this problem instance corresponds to a real rock, and the rover must associate the two on the basis of features that can be detected by its instruments, including its camera. As the rover moves and its camera image changes, the rover must keep track of how those features of the image evolve. This process is uncertain and subject to faults that result in losing track of a rock. In practice, tracking is modeled in the following way:

- In order to perform a measurement on a rock, the rover must be tracking this rock.
- To navigate along a path, it must be tracking one of the rocks that enables following this path. The set of rocks that enable each path is part of the problem definition given to the planner.
- The decision to start tracking a rock must be made before the rover begins to move. Once the rover starts moving, it may keep track of a rock already being tracked or voluntarily stop tracking it, but it cannot acquire a new rock that was not tracked initially.

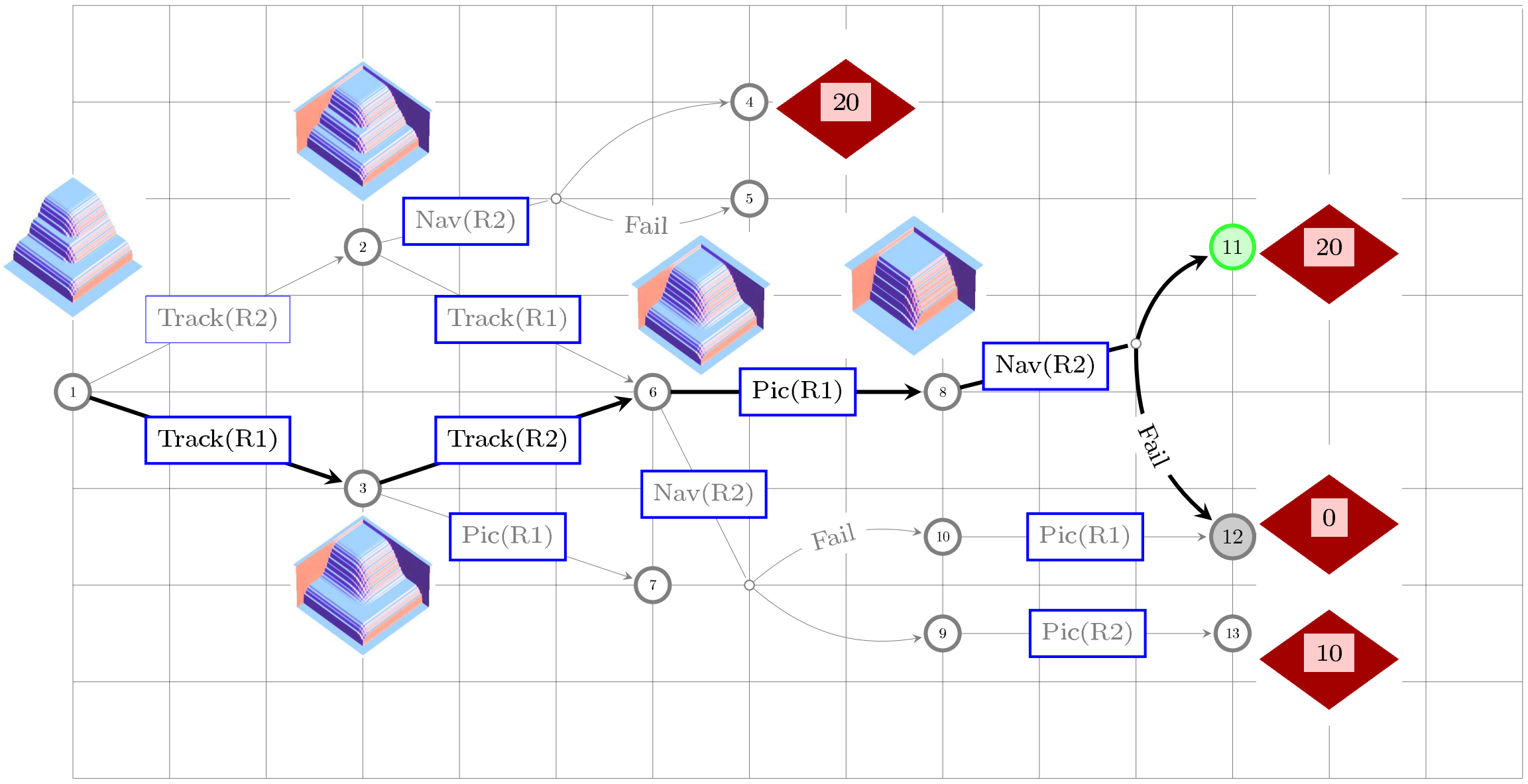

Figure 8: Second iteration of HAO* on the toy problem.

- The rover may randomly lose track of some rocks while navigating along a path. The probability of losing track of a rock depends on the rock and the path followed, it is part of the problem definition given to the planner.
- There is no way to reacquire a rock whose track has been lost, intentionally or by accident.
- The number of rocks tracked strongly influences the duration and resource consumption of navigate actions. The higher the number of rocks tracked, the more costly it is to navigate along a path. This is because the rover has to stop regularly to check and record the aspect of each rock being tracked. This creates an incentive to limit the number of rocks tracked by the rover given the set of goals it has chosen and the path it intends to follow.

So, the rover initially selects a set of rocks to track and tries to keep this set as small as possible given its goals. Once it starts moving, it may lose track of some rocks, and this may cause it to reconsider the set of goals it will pursue and the route to get to the corresponding rocks. It can also purposely stop tracking a rock when this is no longer necessary given the goals that are left to achieve.

Our implementation of HAO* uses the dynamic programming algorithm developed by Feng et al. (2004) and summarized in Section 2.4 in order to perform backups in a hybrid state space, and partitions the continuous state-space associated with a node into piecewise-constant regions. It uses multiple-region updates as described in Section 3.6: an upper bound on the each resource over all expanded regions is computed, and all states included between these bounds and the minimal possible resource levels are updated.

In our experiments, we use the variant of the HAO* algorithm described in Section 3.5, where a parameter $k$ sets the number of times the whole fringe is expanded at each iteration of HAO*; this allows the behavior of the algorithm to be tuned from an exhaustive search to a heuristic search. We used an expansion horizon of $k = 2$ for the simple example in Section 4.1 and a default expansion horizon of $k = 7$ for the larger examples in Section 4.2. Section 4.2.3 describes experiments with different expansion horizons.

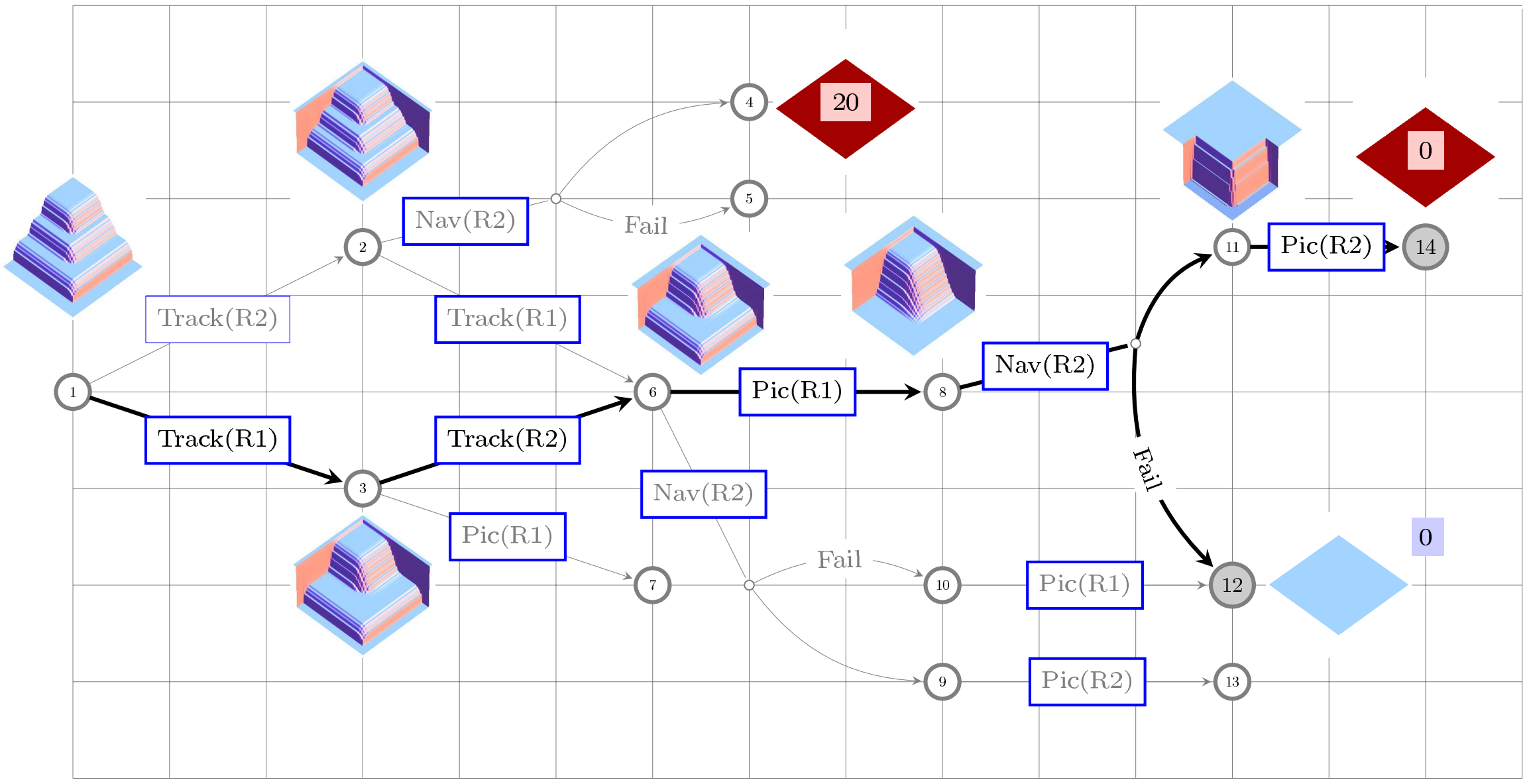

Figure 9: Third iteration of HAO* on the toy problem.

Our implementation of HAO* uses the simple heuristic described in Section 3.4, augmented with a small amount of domain knowledge. The value $H_n(\mathbf{x})$ of a state $(n, \mathbf{x})$ is essentially equal to the sum of the utilities of all goals not yet achieved in $n$. However, if the rover has already moved and a certain rock is not being tracked in state $n$, then all goals requiring this rock to be tracked are not included in the sum. This reflects the fact that, once the rover has moved, it cannot start tracking a rock any more, and thus all goals that require this rock to be tracked are unreachable. The resulting heuristic is admissible (i.e., it never underestimates the value of a state), and it is straightforward to compute. Note that it does not depend on the current resource levels, so that the functions $H_n(\mathbf{x})$ are constant over all values of $\mathbf{x}$.

### 4.1 Example

We begin with a very simple example of the rover planning problem in order to illustrate the steps of the algorithm. We solve this example using the same implementation of HAO* that we use to solve the more realistic examples considered in Section 4.2.

In this example, the targets are two rocks, R1 and R2, positioned at locations L1 and L2, respectively. The rover's initial location is L1, and there is a direct path between L1 and L2. Analyzing rock R1 yields a reward of 10 and analyzing rock R2 yields a reward of 20. The rover's action set is simplified. Notably, it features a single action Pic(Rx) to represents all the steps of analyzing rock Rx, and the "stop tracking" actions have been removed.

Figure 6 shows the optimal value function and the optimal policy found by HAO* for the starting discrete state, and resources ranging over the whole space of possible values. Figures 7, 8 and 9 show the step-by-step process by which HAO* solves this problem. Using an expansion horizon of $k = 2$, HAO* solves this problem in three iterations, as follows:

- **Iteration 1**: As shown in Figure 7, HAO* expands nodes 1, 2 and 3 and computes a heuristic function for the new tip nodes 4, 5, 6 and 7. The backup step yields value function estimates for nodes 1, 2 and 3. HAO* then identifies the best solution graph and a new fringe node 6.

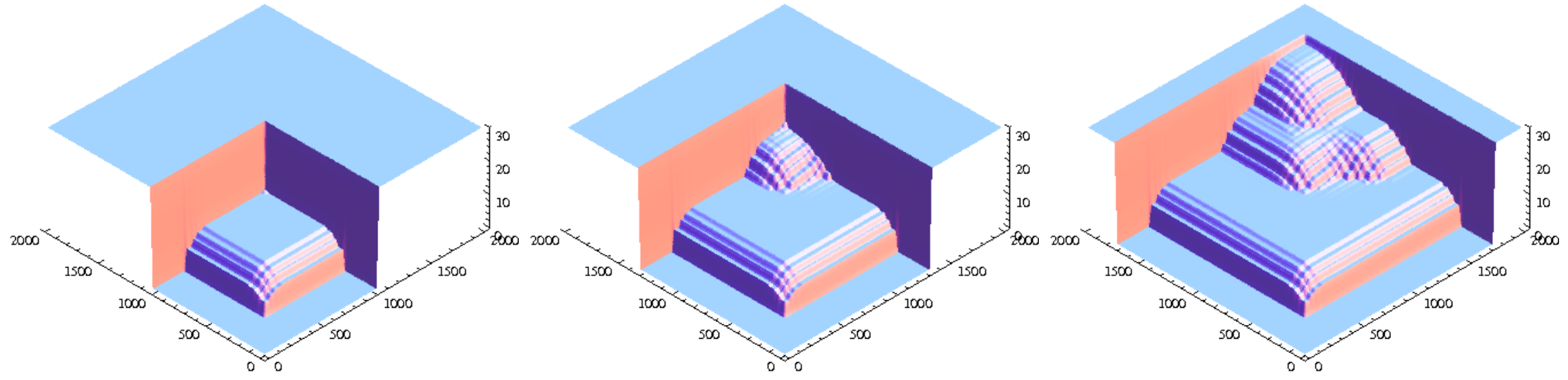

(a) 1012 pieces. (b) 3465pieces. (c) 6122pieces.

Figure 10: Optimal value functions for the initial state of the simple rover problem with increasing initial resource levels (from left to right). The optimal return appears as a three dimensional function carved into the reachable space of the heuristic function.

| problem name | rover loc-ations | paths | goals | fluents | actions | discrete states (approx.) | reachable discrete states | explicit graph | optimal policy | longest branch |
|---|---|---|---|---|---|---|---|---|---|---|
| Rover1 | 7 | 10 | 3 | 30 | 43 | $1.1\ 10^{9}$ | 613 | 234 | 50 | 35 |
| Rover2 | 7 | 11 | 5 | 41 | 56 | $2.2\ 10^{12}$ | 5255 | 1068 | 48 | 35 |
| Rover3 | 9 | 16 | 6 | 49 | 73 | $5.6\ 10^{14}$ | 20393 | 2430 | 43 | 43 |
| Rover4 | 11 | 20 | 6 | 51 | 81 | $2.3\ 10^{15}$ | 22866 | 4321 | 44 | 43 |

Table 3: Size of benchmark rover problems.

- **Iteration 2**: As shown in Figure 8, HAO* expands nodes 6, 8, 9 and 10, starting with previous fringe node 6, and computes heuristic functions for the new tip nodes 11, 12 and 13. The heuristic value for node 12 is zero because, in this state, the rover has lost track of R2 and has already analyzed R1. The backup step improves the accuracy of the value function in several nodes. Node 11 is the only new fringe node since 12 is a terminal node.
- **Iteration 3**: As shown in Figure 9, HAO* expands node 11 and node 14. The search ends after this iteration because there is no more open node in the optimal solution graph.

For comparison, Figure 10 shows how the value function found by HAO* varies with different initial resource levels. In these figures, unreachable states are assigned a large constant heuristic value, so that the value function for reachable states appears as carved in the plateau of the heuristic.

### 4.2 Performance

Now, we describe HAO*'s performance in solving four much larger rover planning problems using the NASA simulation model. The characteristics of these problems are displayed in Tables 3. Columns two to six show the size of the problems in terms of rover locations, paths, and goals. They also show the total number of fluents (Boolean state variables) and actions in each problem. Columns seven to ten report on the size of the discrete state space. The total number of discrete states is two raised to the power of the number of fluents. Although this is a huge state space, only a limited number of states can be reached from the start state, depending on the initial resource levels. The eighth column in Table 3 shows the number of reachable discrete states if the initial time and energy levels are set to their maximum value. (The maximum initial resource levels are based on the scenario of the 2004 IS demo and represent several hours of rover activity.) It shows that simple reachability analysis based on resource availability makes a huge difference. This is partly due to the fact that our planning domain, which is very close to the K9 execution language, does not allow many fluents to be true simultaneously. Columns nine and ten show the number of discrete states in the explicit graph and in the optimal policy. More precisely, the former is the number of nodes created by HAO*, that is, a subset of the reachable discrete states. The number of reachable discrete states, and thus the size of the graph to explore, may seem small compared to other discrete combinatorial problems solved by AI techniques. But each iteration, a continuous approximation of the two-dimensional backup is necessary to evaluate the hybrid state space associated with the graph. Finally, the last column of Table 3 shows the length of the longest branch in the optimal policy when the initial resource levels are set to their maximum value.

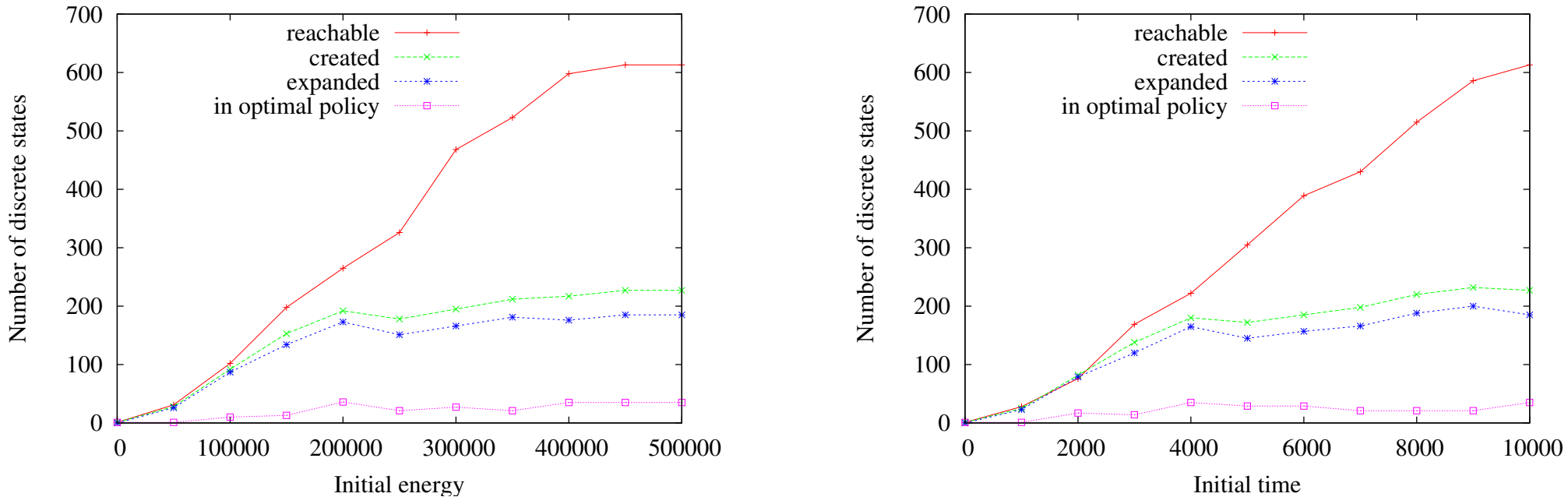

(a) Rover1

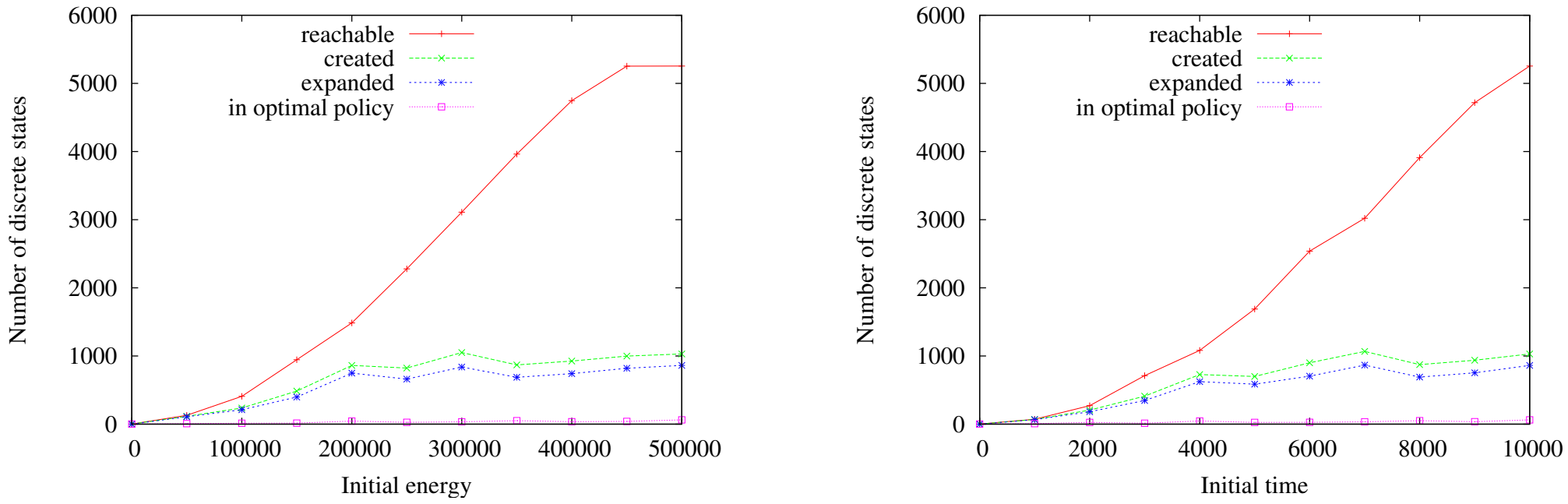

(b) Rover2

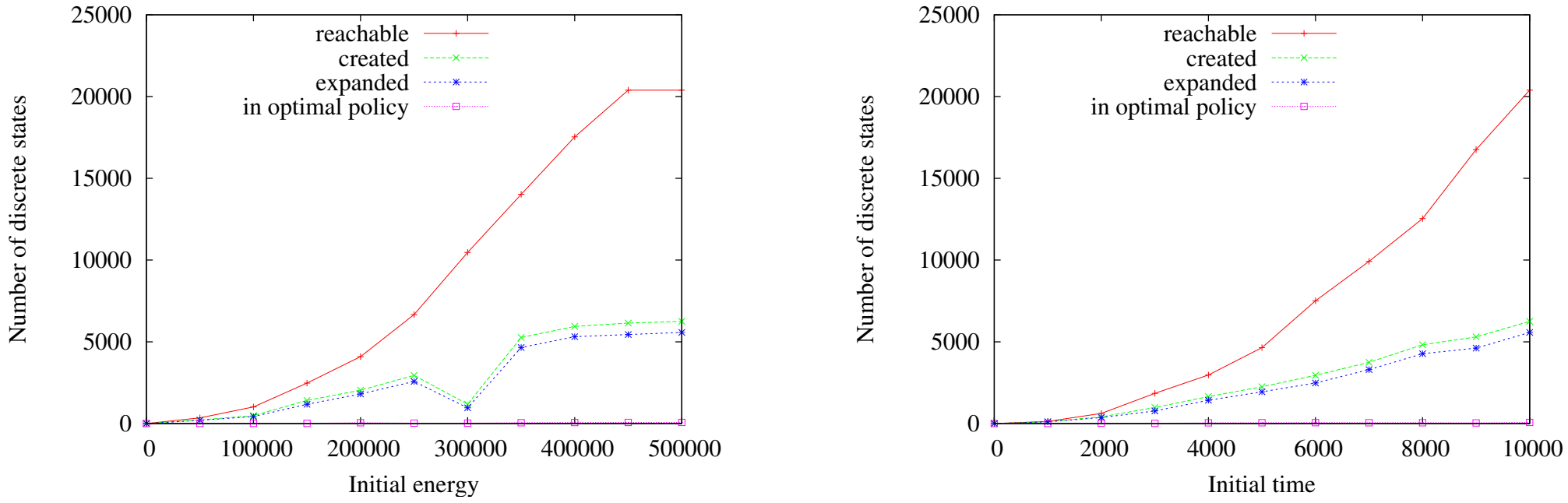

(c) Rover3

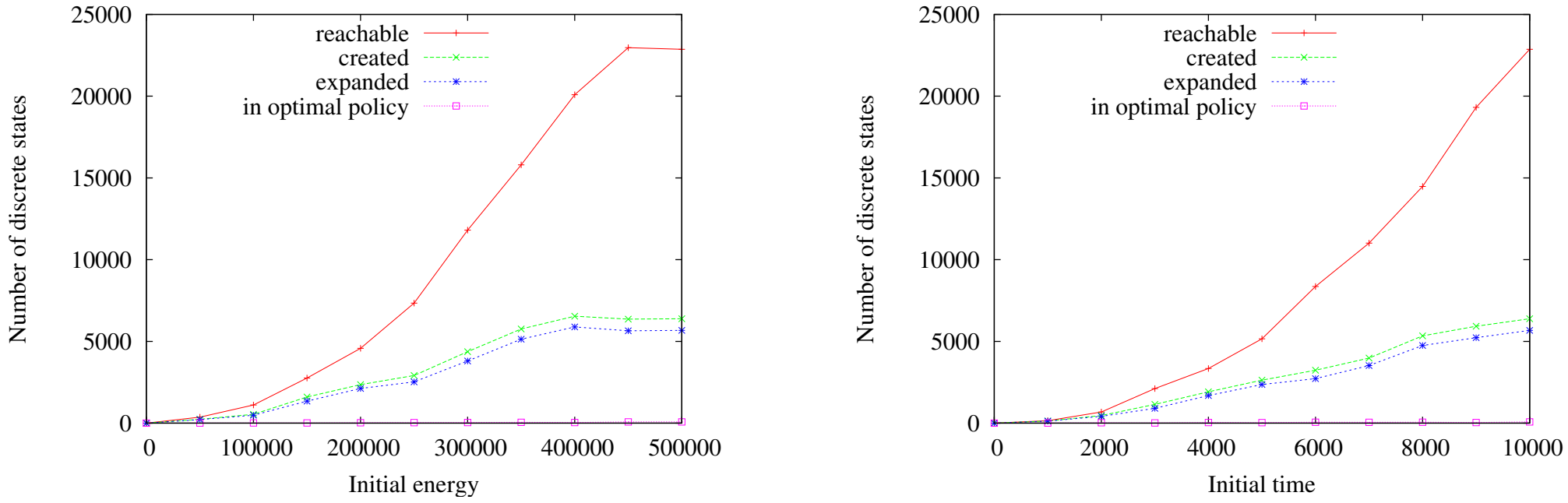

(d) Rover4

Figure 11: Number of nodes created and expanded by HAO* vs. number of reachable discrete states. The graphs in the left column are obtained by fixing the initial time to its maximum value and varying the initial energy. The graphs in the right column are obtained by fixing the initial energy to its maximum value and varying the initial time. Results obtained with $k = 7$.

The largest of the four instances (that is, Rover4) is exactly the problem of the October 2004 IS demo. This is considered a very large rover problem. For example, it is much larger than the problems faced by the MER rovers that never visit more than one rock in a single planning cycle.

#### 4.2.1 Efficiency of Pruning

In a first set of simulations, we try to evaluate the efficiency of heuristic pruning in HAO*, that is, the portion of the discrete search space that is spared from exploration through the use of admissible heuristics. For this purpose, we compare the number of discrete states that are reachable for a given resource level with the number of nodes created and expanded by HAO*. We also consider the number of nodes in the optimal policy found by the algorithm.

Results for the four benchmark problems are presented in Figure 11. These curves are obtained by fixing one resource to its maximum possible value and varying the other from 0 to its maximum. Therefore, they represent problems where mostly one resource is constraining. These result show, notably, that a single resource is enough to constrain the reachability of the state space significantly.

Not surprisingly, problems become larger as the initial resources increase, because more discrete states become reachable. Despite the simplicity of the heuristic used, HAO* is able to by-pass a significant part of the search space. Moreover, the bigger the problem, the more leverage the algorithm can take from the simple heuristic.

These results are quite encouraging, but the number of nodes created and expanded does not always reflect search time. Therefore, we examine the time it takes for HAO* to produce solutions.

#### 4.2.2 Search Time

Figure 12 shows HAO* search time for the same set of experiments. These curves do not exhibit the same monotonicity and, instead, appear to show a significant amount of noise. It is surprising that search time does not always increase with an increase in the initial levels of resource, although the search space is bigger. This shows that search complexity does not depend on the size of the search space alone. Other factors must explain complexity peaks as observed in Figure 12.

Because the number of nodes created and expanded by the algorithm does not contain such noise, the reason for the peaks of computation time must be the time spent in dynamic programming backups. Moreover, search time appears closely related to the complexity of the optimal policy. Figure 13 shows the number of nodes and branches in the policy found by the algorithm, as well as the number of goals pursued by this policy. It shows that: (i) in some cases, increasing the initial resource level eliminates the need for branching and reduces the size of the optimal solution; (ii) the size of the optimal policy and, secondarily, its number of branches, explains most of the peaks in the search time curves. Therefore, the question is: why does a large solution graph induce a long time spent in backups? There are two possible answers to this question: because the backups take longer and/or because more backups are performed. The first explanation is pretty intuitive. When the policy graph contains many branches leading to different combinations of goals, the value functions contain many humps and plateaus, and therefore many pieces, which impacts the complexity of dynamic programming backups. However, we do not have at this time any empirical evidence to confirm this hypothesis. Conversely, we observe that the peak of Figure 12 comes with an increase of the number of backups. More work is required to explain this.

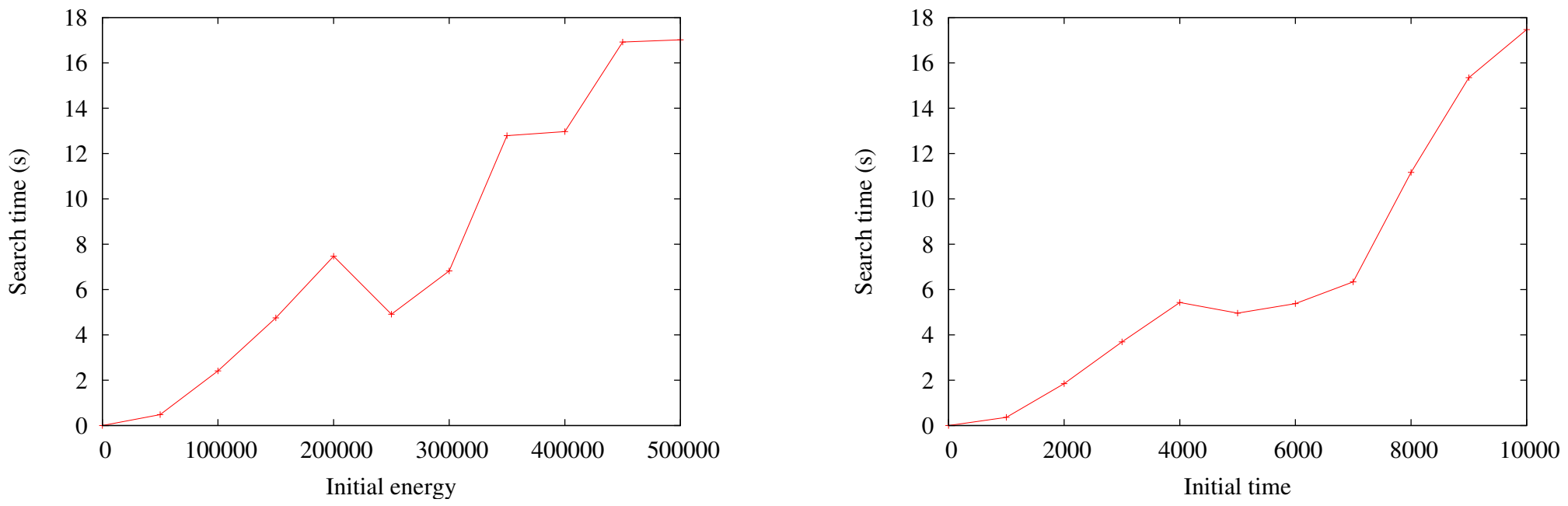

(a) Rover1

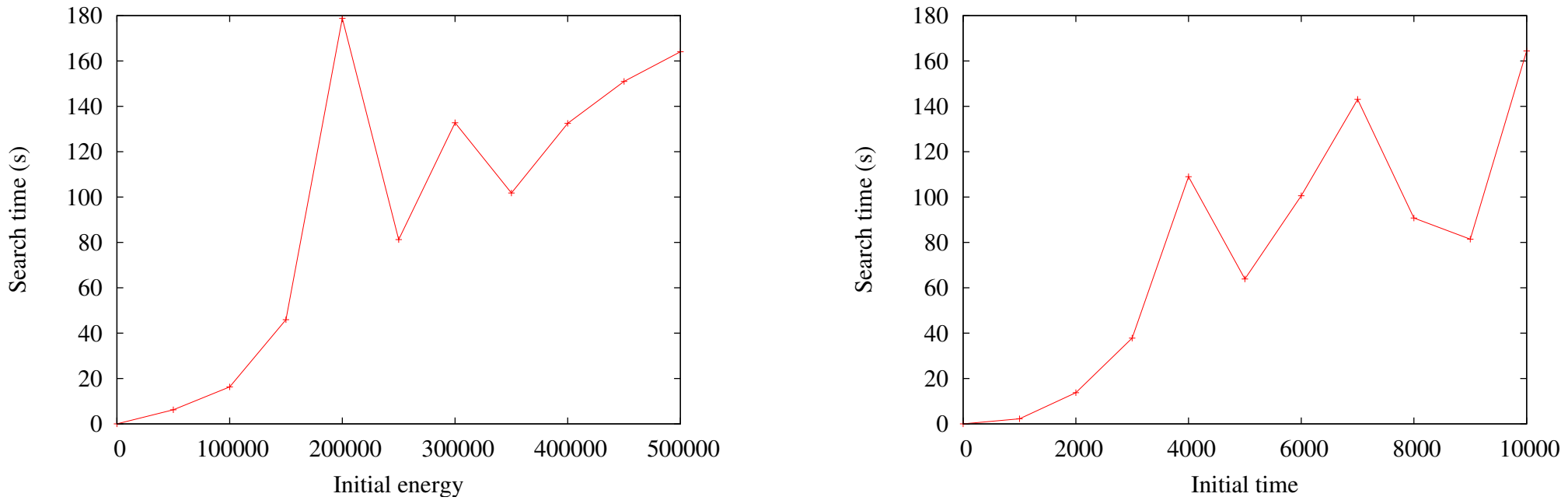

(b) Rover2

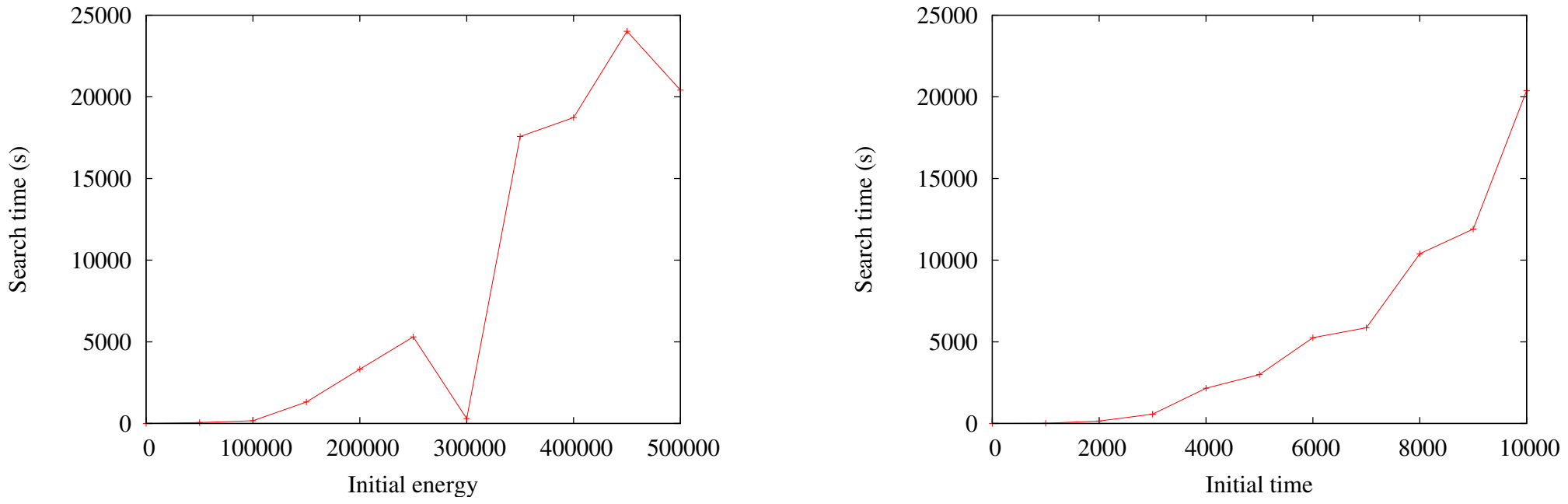

(c) Rover3

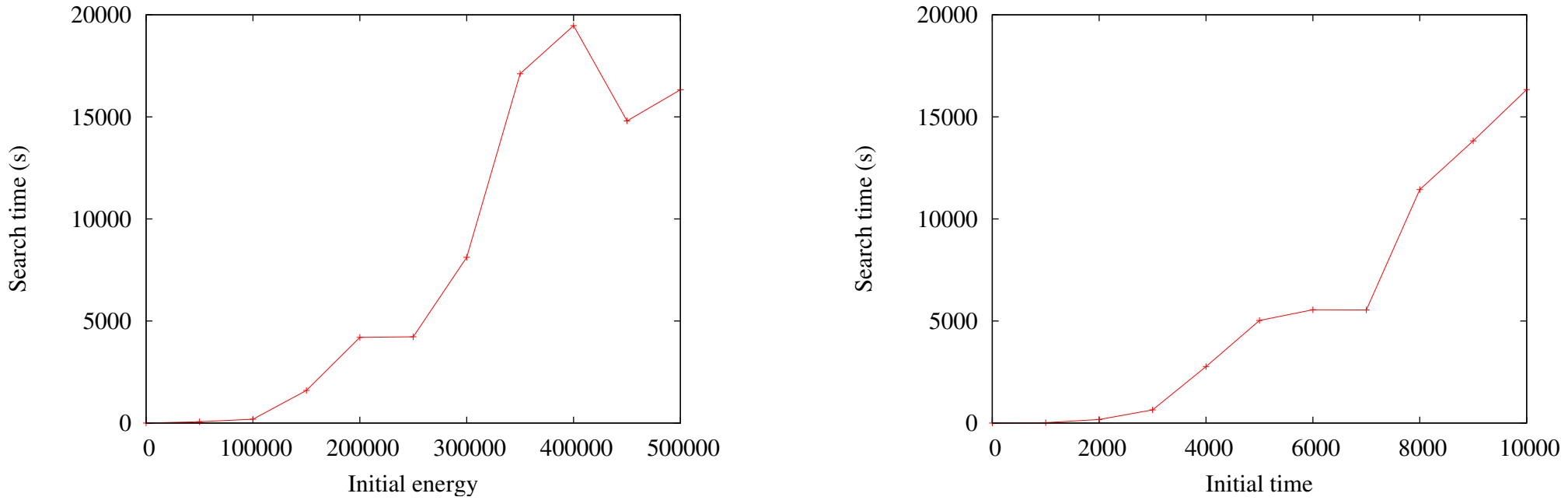

(d) Rover4

Figure 12: HAO* search time. The graphs in the left column are obtained by fixing the initial time to its maximum value, and the graphs in the right column are obtained by fixing the initial energy to its maximum. Results obtained with $k = 7$.

#### 4.2.3 Expansion Horizon

The results of Section 4.2.1 show that HAO* can leverage even a simple admissible heuristic to prune a large portion of the search space. But it does not necessarily follow that HAO* can outperform an "exhaustive search" algorithm that creates a graph of all reachable states, and then executes one pass of dynamic programming in this graph to find the optimal policy. Although HAO* expands a smaller graph than such an exhaustive search, it must evaluate the graph more often. In Section 3.5, we introduced a parameter $k$ for expansion horizon in order to allow adjustment of a trade-off between the time spent expanding nodes and the time spent evaluating nodes. We now study the influence of this parameter on the algorithm.

Figure 14 shows the number of nodes created and expanded by HAO* as a function of the expansion horizon for the four benchmark problem instances. Not surprisingly, the algorithm creates and expands more nodes as the expansion horizon increases. Essentially, it behaves more like an exhaustive search as $k$ is increased. For the two smallest problem instances, and for large enough values of $k$, the number of visited states levels off when the total number of reachable states is reached. For the two largest problem instances, we had to interrupt the experiments once $k$ reached 25 because search time became too long.

Figure 15 shows the effect of the expansion horizon on the search time of HAO*. For the smallest problem instance (Rover1), HAO* does not have a clear advantage over an exhaustive search (with $k > 20$), even though it explores fewer nodes. But for the three larger problem instances, HAO* has a clear advantage. For the Rover2 problem instance, the search time of HAO* levels off after $k = 25$, indicating the limit of reachable states has been reached. However, the duration of such an exhaustive search is several times longer than for HAO* with smaller settings of $k$. The benefits of HAO* are clearer for the two largest problem instances. As $k$ is increased, the algorithm is quickly overwhelmed by the combinatorial explosion in the size of the search space, and simulations eventually need to be interrupted because search time becomes too long. For these same problem instances and smaller settings of $k$, HAO* is able to efficiently find optimal solutions.

Overall, our results show that there is a clear benefit to using admissible heuristics to prune the search space, although the expansion horizon must be adjusted appropriately in order for HAO* to achieve a favorable trade-off between node-expansion time and node-evaluation time.

## 5. Conclusion

We introduced a heuristic search approach to finding optimal conditional plans in domains characterized by continuous state variables that represent limited, consumable resources. The HAO* algorithm is a variant of the AO* algorithm that, to the best of our knowledge, is the first algorithm to deal with all of the following: limited continuous resources, uncertain action outcomes, and over-subscription planning. We tested HAO* in a realistic NASA simulation of a planetary rover, a complex domain of practical importance, and our results demonstrate its effectiveness in solving problems that are too large to be solved by the straightforward application of dynamic programming. It is effective because heuristic search can exploit resource constraints, as well as an admissible heuristic, in order to limit the reachable state space.

In our implementation, the HAO* algorithm is integrated with the dynamic programming algorithm of Feng et al. (2004). However HAO* can be integrated with other dynamic programming algorithms for solving hybrid-state MDPs. The Feng et al. algorithm finds optimal policies under the limiting assumptions that transition probabilities are discrete, and rewards are either piecewise-constant or piecewise-linear. More recently-developed dynamic programming algorithms for hybrid-state MDPs make less restrictive assumptions, and also have the potential to improve computational

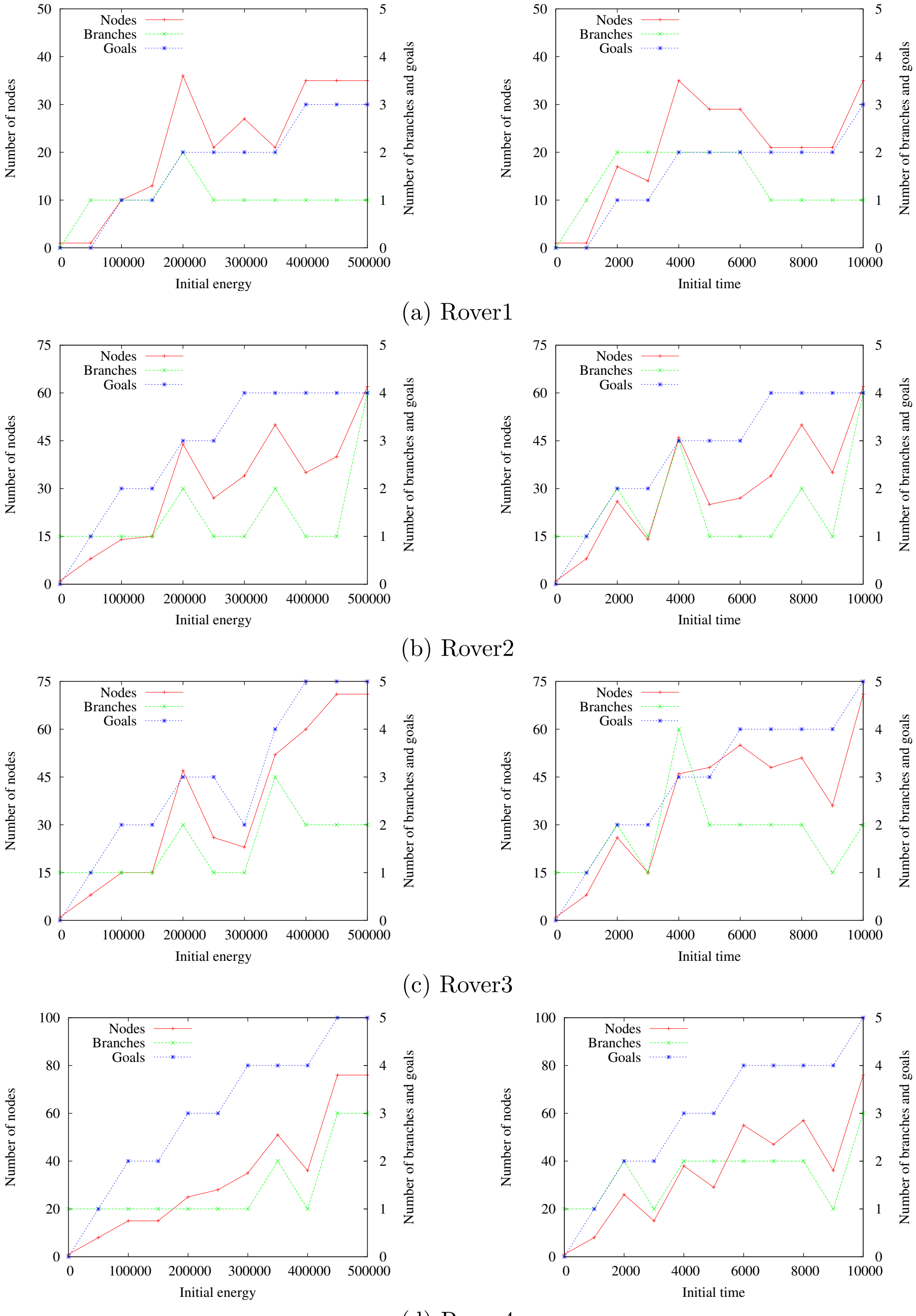

(a) Rover1

(b) Rover2

(c) Rover3

(d) Rover4

Figure 13: Complexity of the optimal policy: number of nodes, branches, and goals in the optimal policy in the same setting as Figure 11.

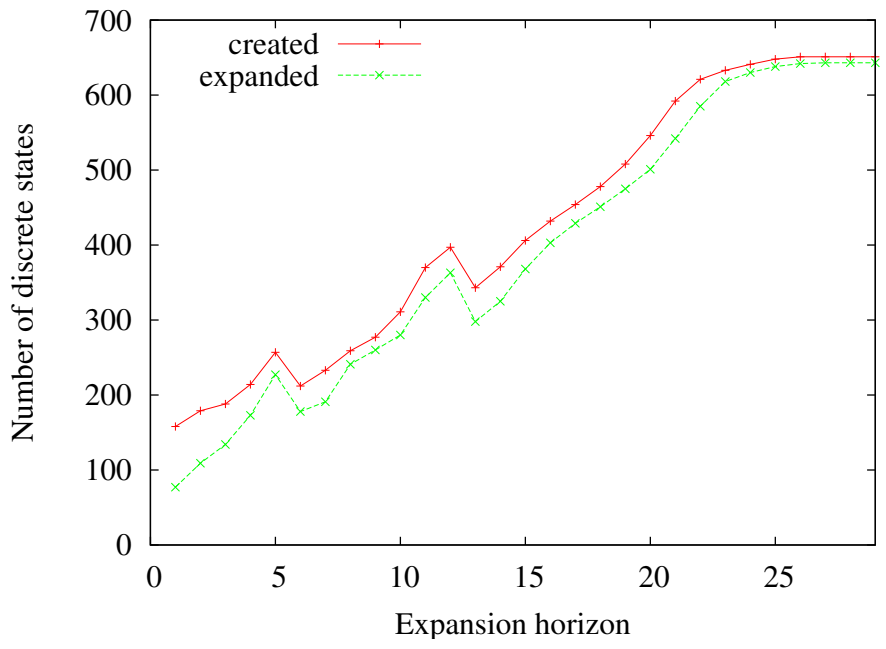

(a) Rover1

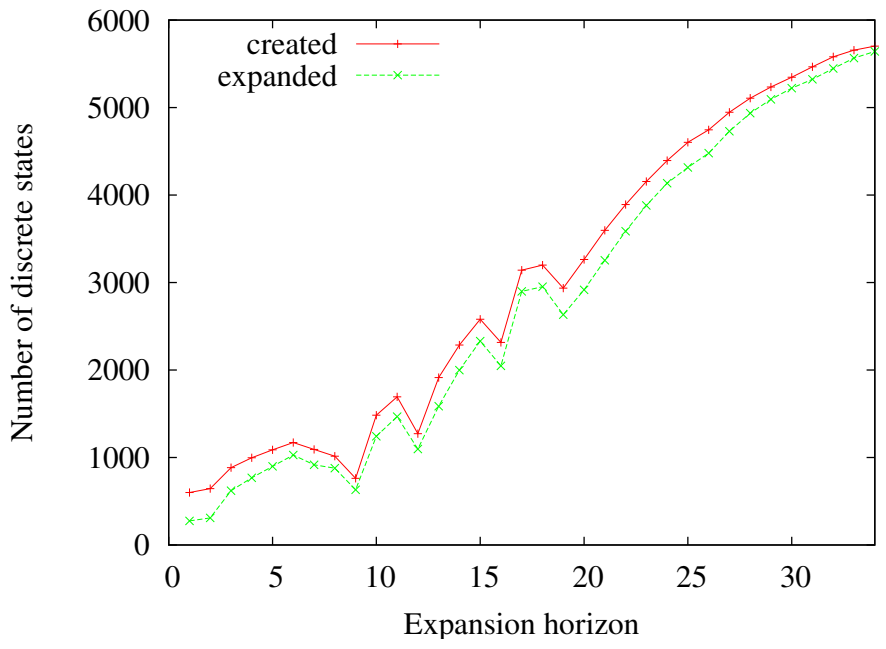

(b) Rover2

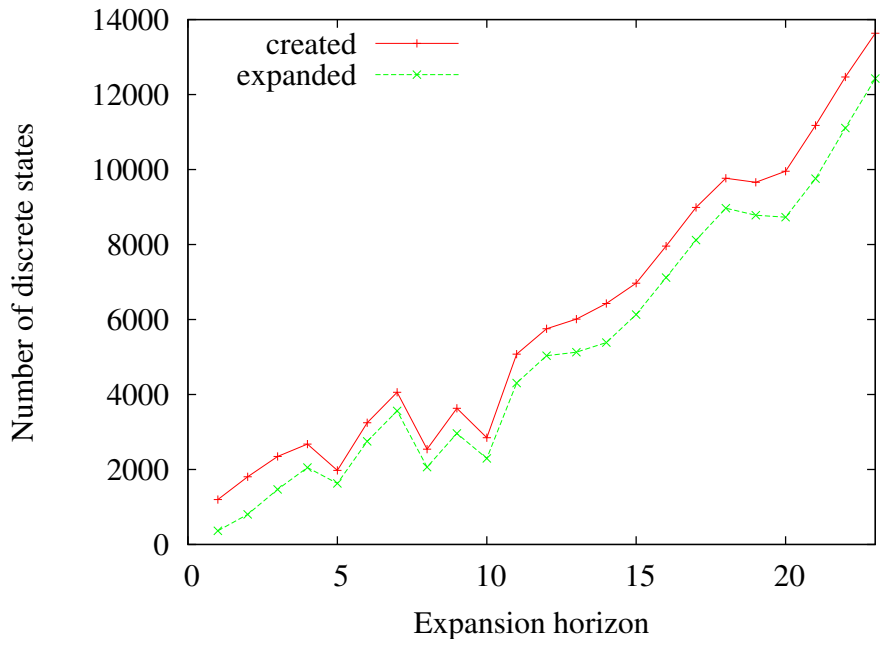

(c) Rover3

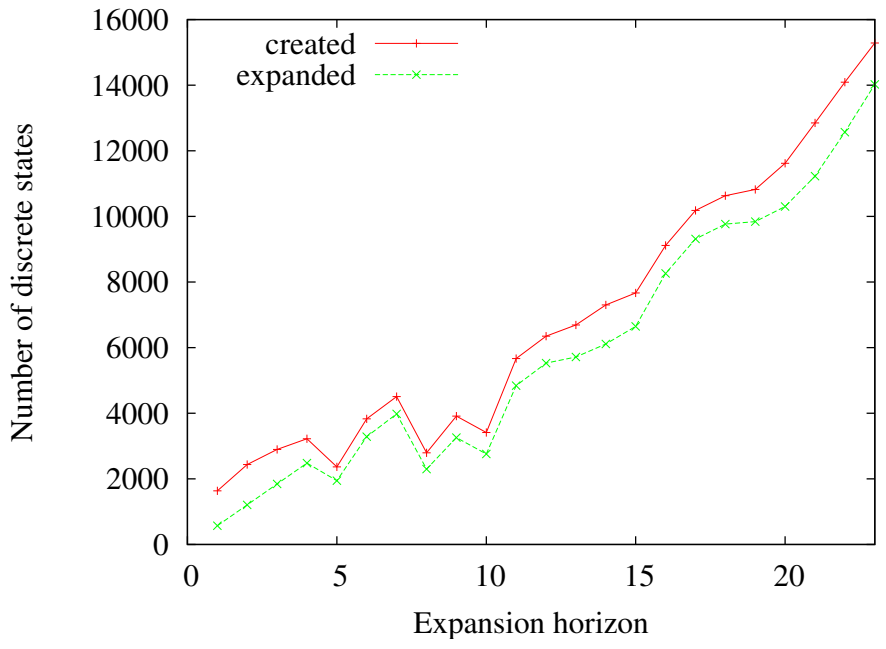

(d) Rover4

Figure 14: Influence of the expansion horizon on the number of nodes visited by the algorithm.

efficiency (Li & Littman, 2005; Marecki et al., 2007). Integrating HAO* with one of these algorithms could improve performance further.

There are several other interesting directions in which this work could be extended. In developing HAO*, we made the assumptions that every action consumes some resource and resources are non-replenishable. Without these assumptions, the same state could be revisited and an optimal plan could have loops as well as branches. Generalizing our approach to allow plans with loops, which seems necessary to handle replenishable resources, requires generalizing the heuristic search algorithm LAO* to solve hybrid MDPs (Hansen & Zilberstein, 2001). Another possible extension is to allow continuous action variables in addition to continuous state variables. Finally, our heuristic search approach could be combined with other approaches to improving scalability, such as hierarchical decomposition (Meuleau & Brafman, 2007). This would allow it to handle the even larger state spaces that result when the number of goals in an over-subscription planning problem is increased.

## Acknowledgments

This work was funded by the NASA Intelligent Systems program, grant NRA2-38169. Eric Hansen was supported in part by a NASA Summer Faculty Fellowship and by funding from the Mississippi Space Grant Consortium. This work was performed while Emmanuel Benazera was working at NASA Ames Research Center and Ronen Brafman was visiting NASA Ames Research Center, both as consultants for the Research Institute for Advanced Computer Science. Ronen Brafman was supported in part by the Lynn and William Frankel Center for Computer Science, the Paul Ivanier Center for Robotics and Production Management, and ISF grant #110707. Nicolas Meuleau is a consultant of Carnegie Mellon University at NASA Ames Research Center.

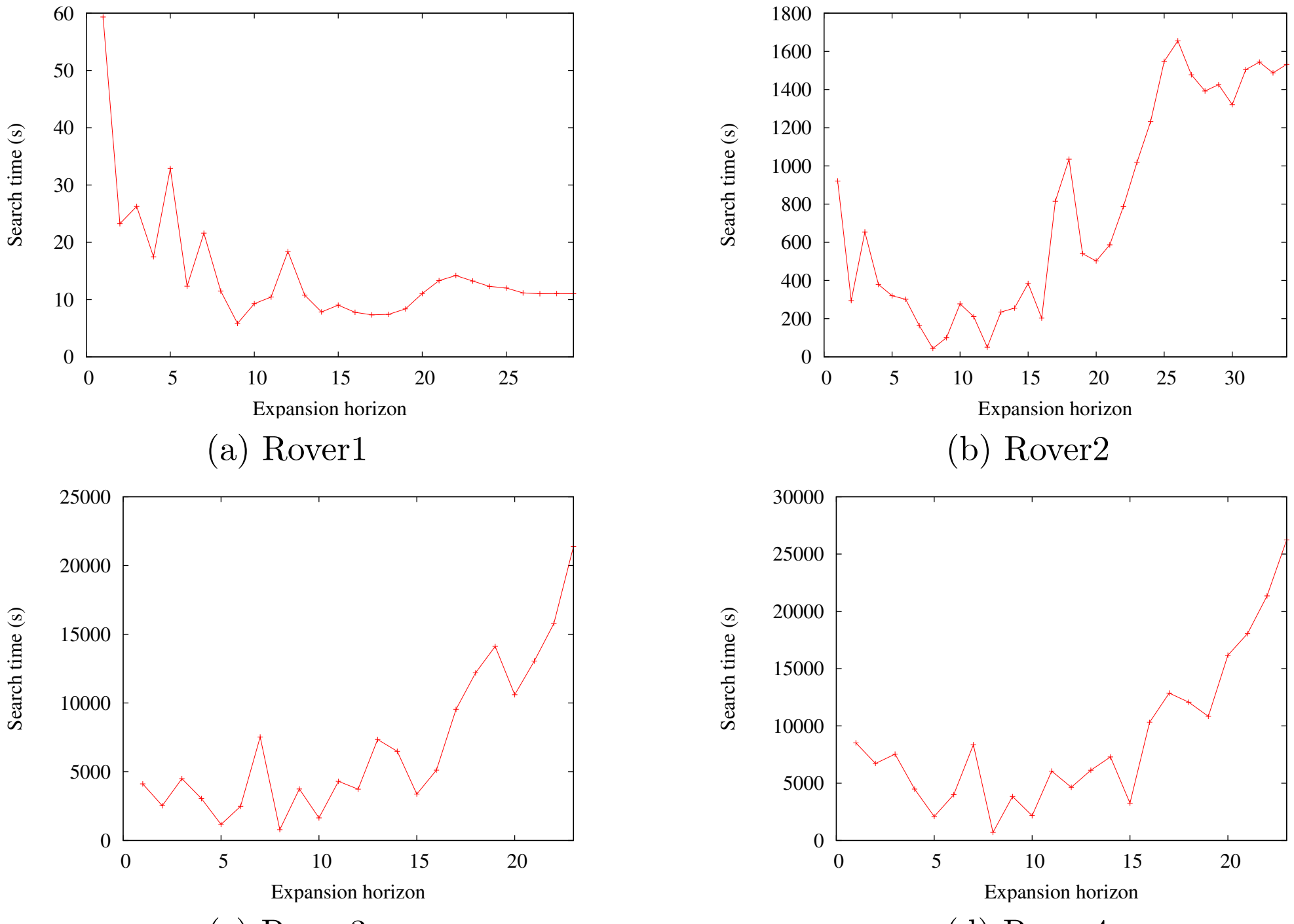

(a) Rover1 (b) Rover2

(c) Rover3 (d) Rover4

Figure 15: Influence of the expansion horizon on overall search time.